%% file: icme2025_template_anonymized.tex
\documentclass[conference]{IEEEtran}
\IEEEoverridecommandlockouts
\usepackage{cite}
\usepackage{amsmath,amssymb,amsfonts}
\usepackage{algorithmic}
\usepackage{graphicx}
\usepackage{textcomp}
\usepackage{xcolor}
\usepackage{multirow}
\usepackage{multicol}
\usepackage{bm}
\usepackage{algorithmic}
\usepackage{algorithm}
\usepackage{tabularx}
\usepackage{booktabs} 
\usepackage{amsmath}
\usepackage{amssymb}
\usepackage{hyperref}
\usepackage{graphicx}
\usepackage{caption}
\title{Enhancing Object Coherence in Layout-to-Image Synthesis}
\author{
    \IEEEauthorblockN{
        Yibin Wang$^{1}$, Changhai Zhou$^{1}$, Honghui Xu$^{2,*}$
    }
    \IEEEauthorblockA{
        $^1$School of Computer Science, Fudan University, Shanghai, China \\
    }
    \IEEEauthorblockA{
        $^2$College of Computer Science and Technology, Zhejiang University of Technology, Hangzhou, China \\
        Email: yibinwang1121@163.com, zhouch23@m.fudan.edu.cn, xhh@zjut.edu.cn
    }
}

\begin{document}
\twocolumn[{

\renewcommand\twocolumn[1][]{#1}%

\begin{center}
    \centering
    \maketitle
    \includegraphics[width=0.8\textwidth]{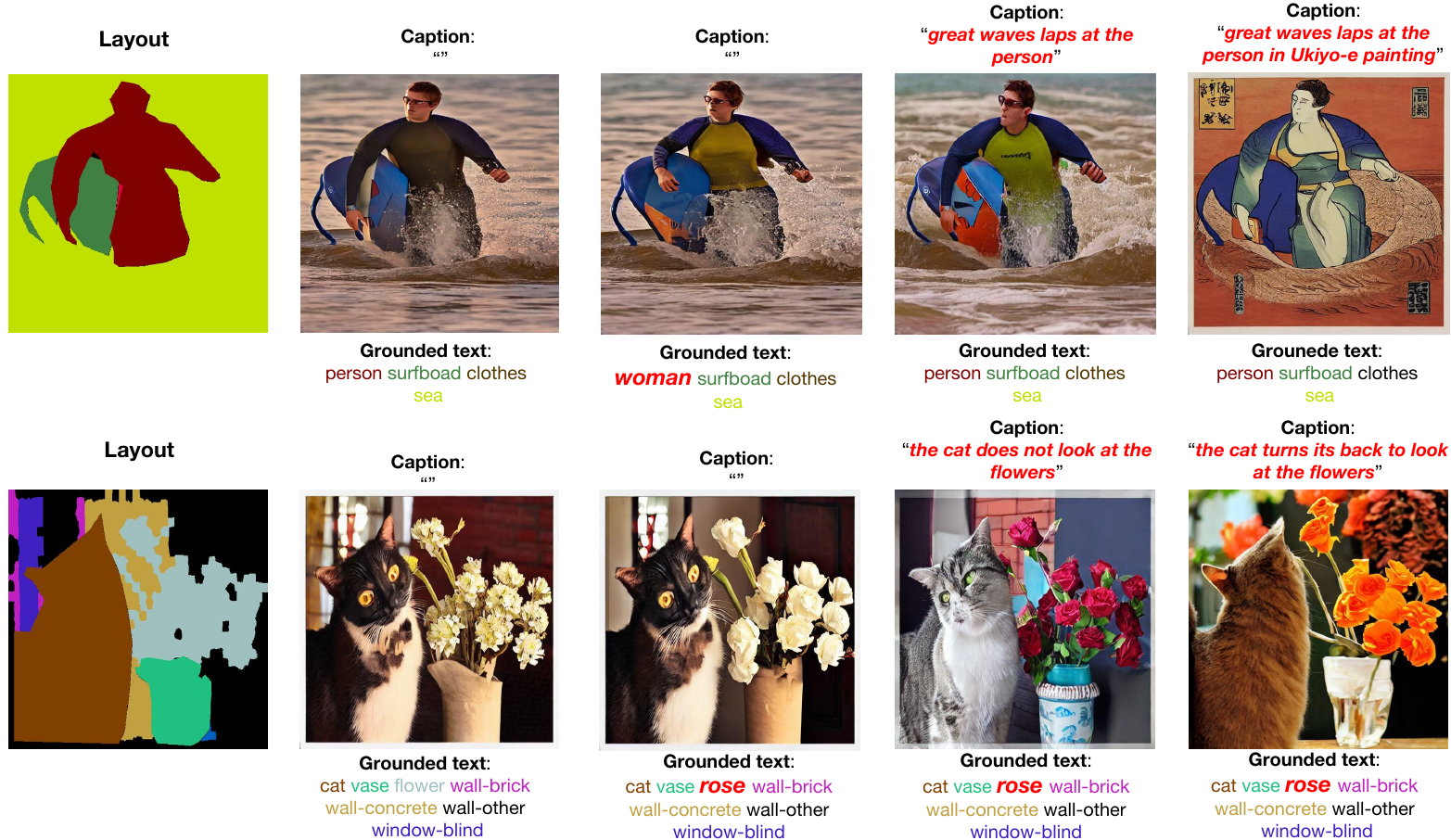}
    \captionof{figure}{Displayed are the LIS outcomes generated using our model. Each instance comprises three inputs: a semantic mask layout, grounded text, and caption. 
    % To exemplify our model's capabilities comprehensively, we present four exemplar outcomes for each layout, showcasing the versatility of our approach by using edited texts and captions.
    }
    \label{fig:display}
\end{center}
}]
\renewcommand{\thefootnote}{}
\footnotetext{$^*$Corresponding author.}

\input{0_abstract}
\input{figs/problem}
\input{figs/model}

\input{1_intro}

\input{2_related}
\input{figs/SFE}

\input{4_method}

\input{5_exp}

\input{7_conclusion}

\bibliographystyle{IEEEbib}
\bibliography{aaai25}
\clearpage
\input{8_appendix}

\end{document}

%% file: 0_abstract.tex
\begin{abstract}
Layout-to-image synthesis aims to generate complex scenes, where users require fine control over the layout of the objects. However, it remains challenging to control the object coherence, including semantic coherence (e.g., the cat looks at the flowers or not) and physical coherence (e.g., the hand and the racket should not be misaligned). In this paper, we propose a novel diffusion model with effective global semantic fusion (GSF) and self-similarity feature enhancement modules to guide the object coherence for this task. For semantic coherence, we argue that the image caption contains rich information for defining the semantic relationship within the objects in the images. Instead of simply employing cross-attention between captions and latent images, which addresses the highly relevant layout restriction and semantic coherence requirement separately and thus leads to unsatisfying results shown in our experiments, we develop GSF to fuse the supervision from the layout restriction and semantic coherence requirement and exploit it to guide the image synthesis process. Moreover, to improve the physical coherence, we develop a Self-similarity Coherence Attention (SCA) module to explicitly integrate local contextual physical coherence relation into each pixel's generation process. Specifically, we adopt a self-similarity map to encode the physical coherence restrictions and employ it to extract coherent features from text embedding. 
Extensive experiments demonstrate the superiority of our proposed method. 
% Through visualization of our self-similarity map, we explore the essence of SCA, revealing that its effectiveness is not only in capturing reliable physical coherence patterns but also in enhancing complex texture generation. 
Code is available at
\href{https://github.com/CodeGoat24/EOCNet}{here}.
% Our model outperforms the previous SOTA methods on FID and DS by relatively 0.9, 3.3\% on COCO-stuff, and 1.1 3.2\% on ADE20K. 
\end{abstract}

%% file: figs/problem.tex
% Use figure* for multi-column figure
\begin{figure*}[ht]

    \centering
    \includegraphics[width=0.9\linewidth]{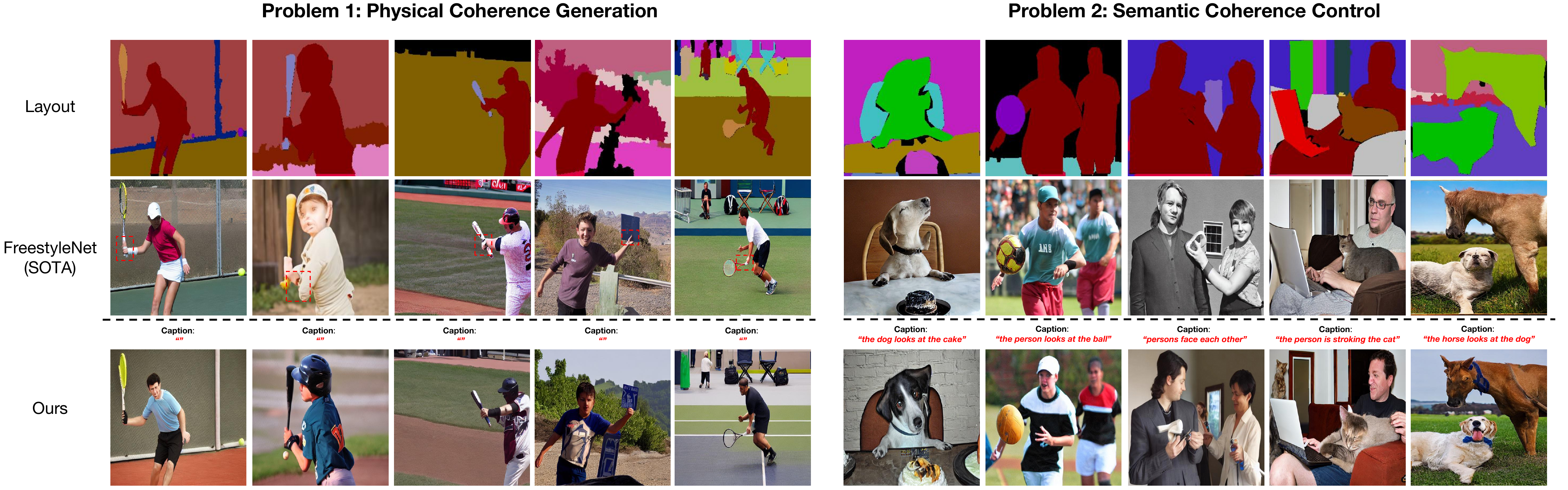}
    \caption{The prevailing challenges of physical coherence and semantic coherence. We omit the grounded text for simplicity and highlight the physical coherence problem using the red outline.}
    \label{fig:problem}
    \vspace{-0.3cm}
\end{figure*}

%% file: figs/model.tex
% Use figure* for multi-column figure
\begin{figure}[t]
    \centering
    \includegraphics[width=1\linewidth]{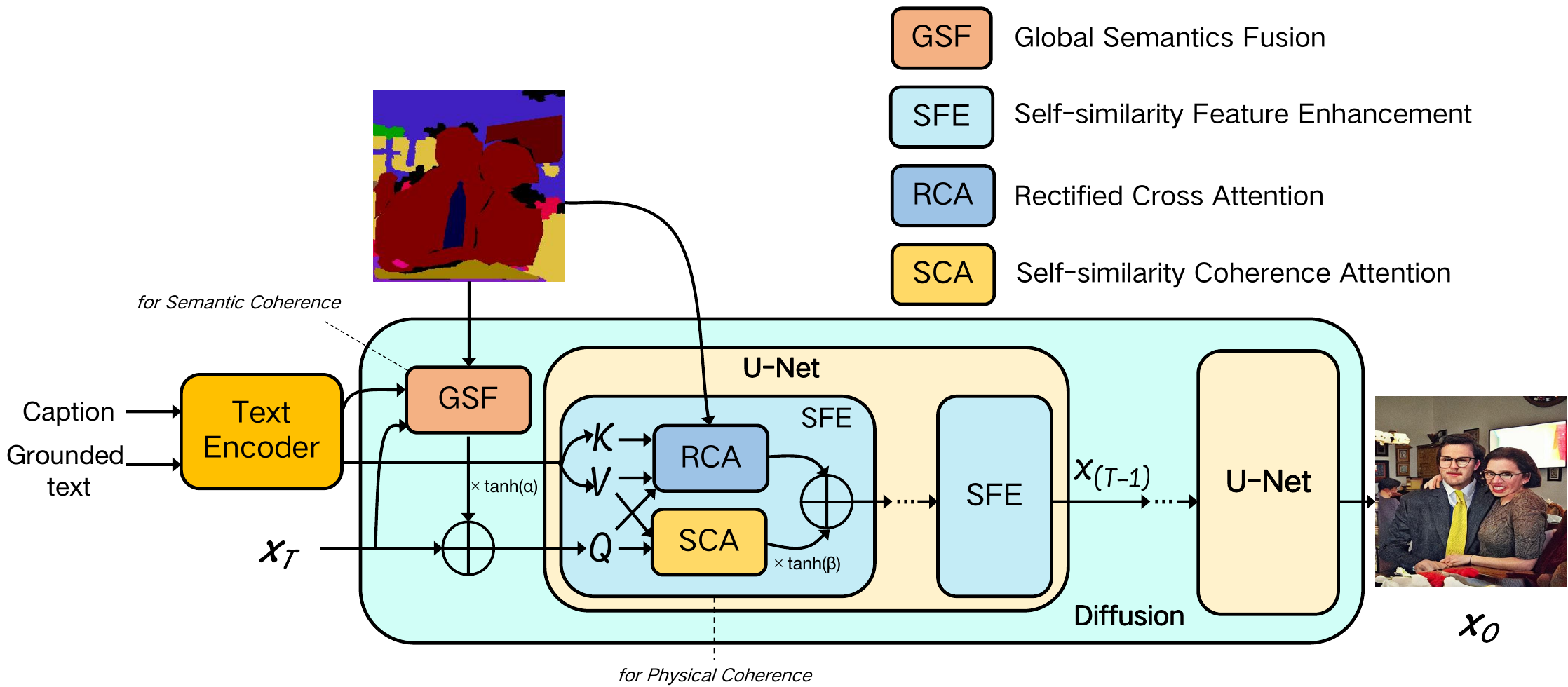}
    \caption{An overview of our EOCNet. 
    % We propose Global Semantics Fusion and Self-similarity Feature Enhancement (SFE) modules to address the semantic coherence and physical coherence problems. 
    % SFE takes the cross-attention's place in the U-net. It leverages the effective synergy between Rectified Cross Attention and our Self-similarity Coherence Attention. 
    %The text encoder and the diffusion model are adopted from Stable Diffusion \cite{rombach2022high} and we omit the autoencoder for simplicity.
    }
    \label{fig:model}
    \vspace{-0.4cm}
\end{figure}

%% file: 1_intro.tex
\section{Introduction}
\label{sec:intro}
Layout-to-image synthesis (LIS) enables the generation of images portraying distinct subjects in diverse scenes.
% In this task, various types of layouts are employed, including bboxes+classes \cite{zheng2023layoutdiffusion, he2023localized}, semantic masks+classes \cite{avrahami2023spatext, SPADE, oasis, freestyleNet}, skeletons \cite{ li2023gligen}, and more. 
% In this work, we focus on the type of semantic masks.
%, using grounded texts to control the object classes and attributes generation (see 2nd and 3rd columns in \cref{fig:display}) and innovatively leveraging the caption to regulate the semantic relation between objects (see last two columns in \cref{fig:display}). 
Earlier studies on 
% mask-based LIS 
employ generative adversarial networks (GANs) \cite{SC-GAN, oasis, CC-FPSE, SPADE, P2PHD,wang2023gan,wang2024learning} . However, these GAN-based approaches often face challenges related to unstable convergence \cite{arjovsky2017towards} and mode collapse \cite{radford2015unsupervised}. Therefore, recent studies \cite{PITI, freestyleNet, avrahami2023spatext,wang2024high,dreamtext} utilize some large-scale pre-trained text-to-image diffusion models like stable diffusion \cite{rombach2022high} to achieve LIS. These pre-trained models exhibit a generative prior with a broad spectrum of semantics, enabling the synthesis of diverse unseen objects. 

However, these methods \cite{freestyleNet, oasis, SPADE, sushko2020you, SC-GAN} still encounter critical challenges concerning object coherence, including semantic coherence (e.g., the cat looks at the flowers or not), and physical coherence (e.g., the hand and the racket should not be misaligned). Taking the recent method, FreestyleNet \cite{freestyleNet}, as an example, which introduced Rectified Cross Attention (RCA) to generate unseen objects from given layouts, it fails to freely control the objects' semantic interactions (Fig. \ref{fig:problem} (right)). Besides, RCA's region-specific token influence neglects implicit physical coherence restrictions between neighboring objects, resulting in unsatisfied image synthesis (Fig. \ref{fig:problem} (left)).
% The qualitative examples presented in  serve as compelling illustrations of these problems. 

We posit that the suboptimal physical generation stems from its inability to capture potential physical coherent restrictions in the layout. Moreover, the failure to control semantic coherence can be attributed to their neglect to incorporate semantic interaction requirements.

In this paper, we propose a novel diffusion model dubbed EOCNet, equipped with our Global Semantic Fusion (GSF) and Self-similarity Feature Enhancement (SFE) modules. These components are designed to address semantic coherence and physical coherence problems, respectively, which will be elaborated as follows.

First, GSF is tailored to excavate the capability for semantic coherence control by integrating semantic coherence requirements into the image generation process. These requirements are defined through our innovatively introduced caption input. It's crucial to note that our caption differs from certain text-to-image studies \cite{rombach2022high, saharia2022photorealistic, kumari2023ablating} that use captions to specify objects' class and position (e.g., "a bag on the desk, next to a vase"), as these are already defined by our grounded text and layout.
Given that the relative positions of objects in the layout inherently restrict their semantic coherence interactions in this task, GSF avoids naive methods (e.g., direct cross-attention \cite{vaswani2017attention} between caption embedding and latent images \cite{rombach2022high}) that address layout restrictions and semantic coherence requirements separately. Instead, GSF effectively merges semantic interaction requirements with spatial restrictions. Specifically, in each time step, GSF first integrates supervision information from the layout map and caption embedding through transformer blocks \cite{vaswani2017attention}. Subsequently, it integrates these amalgamated restriction features into the generated image, providing a comprehensive understanding of both semantic and spatial constraints. This cohesive process effectively guides the image synthesis.

On the other hand,  SFE leverages the effective synergy between RCA and our Self-similarity Coherence Attention (SCA) to enhance the physical coherence generation. To be precise, it extends the superiority of RCA \cite{freestyleNet} and remedies its drawback through our ingeniously designed SCA.
To explicitly integrate inherent physical coherence restrictions into each pixel's generation process, SCA revolutionizes the way each pixel interacts with its neighboring features, effectively capturing local contextual physical coherence relations.
It encodes these relations into vectors, generating similarity maps that encapsulate rich coherence relationships between each pixel and its contextual features. These maps are then utilized to extract features with potential physical coherence restrictions from text embeddings. This process enhances the generator's ability to comprehend intricate physical coherence relationships in the layout, contributing to optimal physical coherence generation. 
To form a synergy between RCA and SCA, their outputs are fused, allowing the model to leverage SCA's fine-grained understanding of local context while retaining RCA's proficiency in generating previously unseen objects \cite{freestyleNet}. 
% This integration enhances the model's overall ability to generate images with improved physical coherence.
%By doing so, SCA adeptly models the local context, enhancing the model's ability to comprehend the intricate physical coherence relationships in the layout.

% Finally, it is notable to highlight that SFE leverages the effective synergy between RCA and SCA. This empowers the model to leverage SCA's fine-grained understanding of local context while retaining RCA's superiority in generating previously unseen objects \cite{freestyleNet}. 
We evaluate our model on COCO-Stuff \cite{caesar2018coco} and ADE20K \cite{zhou2017scene}, and our model outperforms the previous SOTA methods on FID and DS by relatively 0.9, 3.3\% on COCO-stuff, and 1.1, 3.2\% on ADE20K.
The visualized self-similarity maps in SCA validate its innate proficiency in discerning dependable physical coherence patterns. Notably, it also underscores SCA's surprising efficacy in enhancing the generation of complex textures. 
This multifaceted superiority of SCA propels the model's capability to yield more high-fidelity images.

Our main contributions are threefold. (1)We address the coherence problems in LIS. Specifically, we excavate the capability for semantic coherence control and improve the physical coherence generation, achieving greater controllability and higher generation quality; (2) By exploring the essence of our SCA, we reveal its effectiveness not only in capturing reliable physical coherence patterns but also in enhancing complex texture generation; (3) Our model outperforms the previous SOTA methods on FID and DS by relatively 0.9, 3.3\% on COCO-stuff and 1.1, 3.2\% on ADE20K.

% \begin{itemize}
% 	\item We address the coherence problems in LIS. Specifically, we excavate the capability for semantic coherence control and improve the physical coherence generation, achieving greater controllability and higher generation quality.
% 	\item By exploring the essence of our SCA, we reveal its effectiveness not only in capturing reliable physical coherence patterns but also in enhancing complex texture generation. 
% 	\item Our model outperforms the previous SOTA methods on FID and DS by relatively 0.9, 3.3\% on COCO-stuff and 1.1, 3.2\% on ADE20K
% \end{itemize}

%% file: 2_related.tex
\section{Related Work}
\label{sec:related}
% \subsection{Semantic Layout Image Synthesis}
This task aims to generate multiple objects in diverse scenes using given semantic layouts \cite{avrahami2023spatext, alaniz2022semantic, P2P, CC-FPSE, lv2022semantic, SPADE, shi2022retrieval, oasis, tan2022semantic, P2PHD, wang2022semantic, SC-GAN, zhu2022label}. Existing approaches have been explored in the field of semantic image synthesis. For instance, Pix2Pix \cite{P2P} employs an encoder-decoder generator with a PatchGAN discriminator. 
% Pix2PixHD \cite{P2PHD} builds on Pix2Pix by introducing coarse-to-fine and multi-scale network architectures to handle high-resolution image synthesis.
SPADE \cite{SPADE} modulates the activations in normalization layers using affine parameters predicted from input semantic maps. 
% On the other hand, CC-FPSE \cite{CC-FPSE} and SC-GAN \cite{SC-GAN} adopt a different approach by learning to generate convolutional kernels and semantic vectors from the semantic maps to condition the image generation process. 
Additionally, OASIS \cite{oasis} introduces a segmentation-based discriminator to improve image alignment with the input label maps. Despite their impressive performance, these methods suffer from challenges such as unstable convergence and model collapse, mainly due to their reliance on GANs. These issues can result in artifacts and a lack of coherence in the generated images, limiting their overall quality. 
Recognizing GANs' inherent limitations, PITI \cite{PITI} and ControlNet \cite{controlnet} pioneeringly adopt a pre-trained diffusion model to enhance image-to-image translation. Following this, FreestyleNet \cite{freestyleNet} introduces a freestyle LIS framework, emphasizing controllability. 
% This framework empowers the generation of semantics beyond the confines of pre-defined semantic categories present in the training dataset. 
% Notably, SpaText \cite{avrahami2023spatext} proposes a CLIP-based spatio-textual representation, enabling the specification of each region using free-form text. This approach effectively achieves free-form textual scene control.

However, they are still facing critical challenges concerning object coherence. In this paper, we present our framework to tackle both semantic and physical coherence in the LIS task. Our primary goal is to introduce a novel method that achieves superior generation quality and enhanced controllability compared to the mentioned approaches.

% \subsection{Diffusion Model}
% Diffusion models \cite{bao2022analytic, ho2020denoising, nichol2021improved, rombach2022high, yang2022diffusion} are emerging as promising generative models, setting new benchmarks in image generation across various domains \cite{ulhaq2022efficient, yang2022diffusion, careil2023few, dong2023cvsformer}, including class-conditional \cite{dhariwal2021diffusion, zheng2022entropy}, text-to-image \cite{rombach2022high, saharia2022photorealistic, kumari2023ablating}, and image-to-image translation tasks \cite{kawar2022denoising, lugmayr2022repaint, saharia2022image, jiang2023masked, park2023lanit}. Notably, ADM-G \cite{dhariwal2021diffusion} explores the classifier guidance, using the classifier's gradient on noisy images during sampling. Further, Ho et al. \cite{ho2022classifier} introduce a classifier-free approach by interpolating between predictions of a diffusion model with and without conditional input. For efficient training and sampling, LDM compresses images to a smaller resolution and employs denoising training in the latent space.

%% file: figs/SFE.tex
% Use figure* for multi-column figure
\begin{figure}[t]
    \centering
    \includegraphics[width=0.9\linewidth]{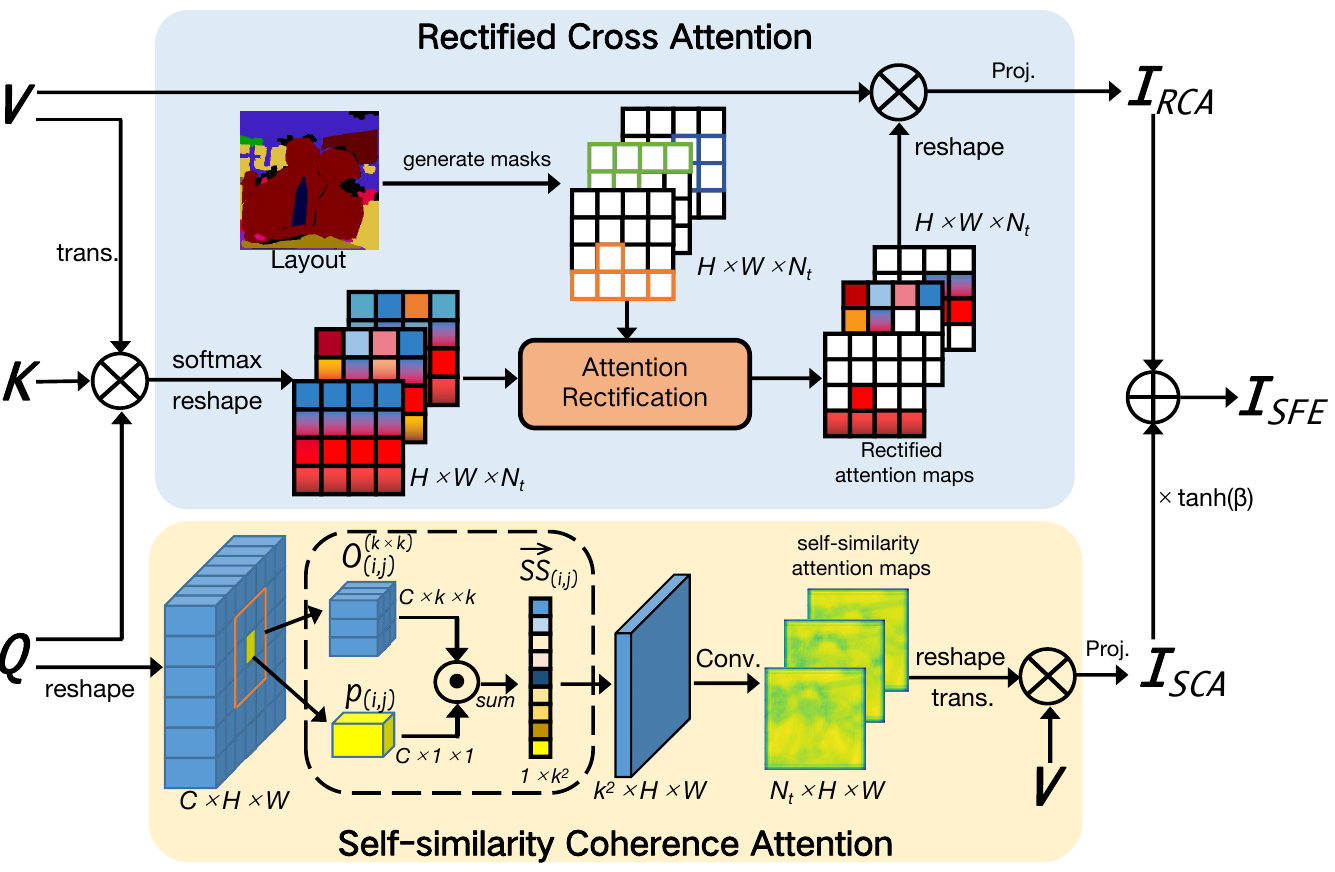}
    \caption{The pipeline of our SFE module.}
    \label{fig:SFE}
    \vspace{-0.5cm}
\end{figure}

%% file: 4_method.tex
\section{Method}
\label{sec:method}
% In this section, we will begin by providing an overview and the training pipeline of our method. Following that, we will provide a detailed explanation of the Global Semantics Fusion (GSF) module. Subsequently, we will delve into the intricacies of the Self-similarity Feature Enhancement (SFE) module, shedding light on its fundamental role in fostering a synergistic interplay between Self-similarity Coherence Attention (SCA) and Rectified Cross Attention (RCA).

%Subsequently, we provide detailed insights into our approach, focusing on two key aspects: 1) The operation of GSF in conditioning the image through the fusion of global spatial restriction and semantic requirement. 2) The synergistic interplay between SCA and RCA within the SFE module, enhances physical coherence generation while preserving the ability to generate previously unseen objects.

\subsection{Overview}
In this work, we aim to tackle both semantic coherence and physical coherence issues in the LIS task. As illustrated in Fig. \ref{fig:model}, our approach incorporates the GSF to integrate semantic coherence requirements into the synthesis process and employs SFE to enhance the generation of physical coherence.

% Given the initial noise image $\textbf{\textit{z}}_{T}$ perturbed by Gaussian noise $\bm{\varepsilon}$, grounded text, caption, and semantic mask, we first stack all concepts from the grounded text. Then we feed the stacked concepts and the caption into the text encoder CLIP \cite{radford2021learning}, outputting text embedding $\textbf{\textit{y}}_{g} \in \mathbb{R}^{C_{t} \times N_{t}}$ and caption embedding $\textbf{\textit{y}}_{e} \in \mathbb{R}^{C_{t} \times N_{t}}$, where $C_{t}$ and $N_{t}$ denote the amount of channels and tokens respectively. After that, we resize the single-channel semantic layout to align with the spatial dimensions of the noise images $\textbf{\textit{z}}_{T} \in \mathbb{R}^{C_{x} \times H \times W}$ using interpolation, yielding the layout embedding $\textbf{\textit{L}} \in \mathbb{R}^{1 \times H \times W}$, where \textit{H} and \textit{W} denote the height and width. 
% Based on these, our approach synthesizes the image $\textbf{\textit{z}}_{0}$ over \textit{T} time steps in a progressive and evolving manner.

% \subsection{Training Process}
Specifically, in each time step \textit{t}, we first fuses supervision information from the layout embedding \textbf{\textit{L}} and caption embedding $\textbf{\textit{y}}_{e}$ through GSF. Then, these amalgamated supervision features are infused into the noise image $\textbf{\textit{x}}_{t}$ to steer the image synthesis effectively. After that, the constricted image $\textbf{\textit{x}}_{t}^{'}$ will be input to the U-Net $\varepsilon_{\theta}$  whose cross-attention layer is replaced with our SFE. Illustrated in Fig. \ref{fig:SFE}, SFE facilitates a dynamic interplay between RCA and SCA
Precisely, RCA restricts each token embedding to influence only its corresponding area, while SCA encodes local potential coherence relations of each pixel into self-similarity maps, thereby enhancing the generation of physical coherence. 
Ultimately, the output features of SCA and RCA are amalgamated, yielding the final output of SFE.

The objective loss for our fine-tuning is formulated as follows:
\begin{equation}
    	\mathcal{L}(\theta) = \mathbb{E}_{\textbf{\textit{z}},\textbf{\textit{L}},\textbf{\textit{y}}_{g},\textbf{\textit{y}}_{e},\bm{\varepsilon} \sim N(0,1), t}  \parallel \bm{\varepsilon} - \varepsilon_{\theta}(\textbf{\textit{z}}_{t}, t, \textbf{\textit{L}}, \textbf{\textit{y}}_{g}, \textbf{\textit{y}}_{e} \parallel_{2}^{2}. \nonumber
\end{equation}

\subsection{Global Semantics Fusion}
% We propose GSF module to model the spatial position and semantic interaction relationships between objects in the image, thus providing a latent representation of the entire image with respect to both layout and semantic relationships. Specifically, 
GSF is tailored to achieve semantic coherence control by simultaneously integrating semantic coherence requirements and layout restriction into the image generation process. Specifically,
It employs several sequential layers of classical self- and cross-attention blocks \cite{vaswani2017attention,wang2024residual} to effectively integrate the supervision information from layout embedding $\textbf{\textit{L}} \in \mathbb{R}^{1 \times H \times W}$ and caption embedding $\textbf{\textit{y}}_{e} \in \mathbb{R}^{C_{t} \times N_{t}}$. First, it concatenate $\textbf{\textit{z}}_{t}$ and \textbf{\textit{L}}, followed by employing a linear mapping to obtain $\textbf{\textit{L}}^{'} \in \mathbb{R}^{C \times H \times W}$. Then, the integration process within layer \textit{l} $\in [0, g]$ can be formulated as follows.
\begin{align}
	\textbf{\textit{F}}_{l} &= \textup{SelfAtt}(\textbf{\textit{O}}_{l-1}, \textbf{\textit{O}}_{l-1}, \textbf{\textit{O}}_{l-1}) + \textbf{\textit{O}}_{l-1},\\
	\textbf{\textit{F}}_{l}^{'} &= \textup{CrossAtt}(\textbf{\textit{F}}_{l}, \textbf{\textit{y}}_{e}, \textbf{\textit{y}}_{e}) + \textbf{\textit{F}}_{l},\\
        \textbf{\textit{O}}_{l} &= \textbf{\textit{F}}_{l}^{'}\textbf{\textit{W}}_{l},
\end{align}
where $\textbf{\textit{O}}_{l}, \textbf{\textit{F}}_{l}, \textbf{\textit{F}}_{l}^{'} \in \mathbb{R}^{C \times H \times W}$, $\textbf{\textit{O}}_{0} = \textbf{\textit{L}}^{'}$, $\textbf{\textit{W}}_{l}$ representing the parameters of the feedforward layer and $\textbf{\textit{O}}_{g}$ is the output of GSF. 

To infuse amalgamated supervision features into the image, a straightforward approach is to directly incorporate $\textbf{\textit{O}}_{g}$ into the image features $\textbf{\textit{z}}_{t}$ \cite{zheng2023layoutdiffusion}. 
This process can be defined as
\begin{align}
	\textbf{\textit{z}}_{t}^{'} = \textbf{\textit{z}}_{t} + \textup{tanh}(\alpha) * \textbf{\textit{O}}_{g},
\end{align}
where $\alpha$ represents a learnable parameter initialized with a value of 0.

\subsection{Self-similarity Feature Enhancement}
We propose SFE that fosters a dynamic interplay between RCA and SCA to substitute the cross-attention layer within the U-Net. 
In this section, we will first introduce RCA's effectiveness in restricting each token's influence over pixels within the specific region. Then, we will delve into the details of SCA on guiding each pixel to engage with its neighboring features. Finally, the fusion mode of the output features from SCA and RCA is elaborated.

%thereby effectively capturing local contextual physical coherence relations, and 3) fostering a dynamic interplay between RCA and SCA.

\input{figs/performance}

\subsubsection{Rectified Cross Attention}
We utilize the RCA to ensure that each object (defined by grounded text) is generated within the region defined by the layout. It compels each text token to influence the designated region by rectifying the computed attention maps. Specifically, given the image feature maps and text embedding $\textbf{\textit{y}}_{g}$, SFE computes image queries \textbf{\textit{Q}} $\in \mathbb{R}^{HW \times C}$, text keys \textbf{\textit{K}}, and text values \textbf{\textit{V}} $\in \mathbb{R}^{N_{t} \times C}$ through three separate projection layers. 
The cross-attention maps \textbf{\textit{A}} $\in\mathbb{R}^{(H\times W)\times N_{t}}$ is computed as 
$\textbf{\textit{A}} = \textup{reshape}(\textbf{\textit{Q}}\cdot\textbf{\textit{K}}^{\mathrm{T}}/\sqrt{d})$, where \textit{d} is the the dimension of the queries and keys.

Then, attention map $\textbf{\textit{A}}^{k}$ in $\textbf{\textit{A}}$ are rectified by applying its corresponding layout $\textbf{\textit{l}}^{k}$ derived from \textbf{\textit{L}}:
\begin{align}
\textbf{\textit{A}}^{k}=\begin{cases}\phantom{-}a_{ij}^{k},&l_{ij}^{k}=1,\\-inf,&l_{ij}^{k}=0,\end{cases} \label{equ:rectify}
\end{align}
where $a_{ij}^{k}$ and $l_{ij}^{k}$ represents the weight of the \textit{k}-th token's attention map and the value of $\textbf{\textit{l}}^{k}$, respectively, at the position (\textit{i}, \textit{j}). After that, we obtain the rectified attention maps $\hat{\textbf{\textit{A}}}$ and the output of RCA layer is defined as $\textbf{\textit{I}}_{RCA} \in \mathbb{R}^{HW \times C_{x}} =$Softmax$(\hat{\textbf{\textit{A}}})\textbf{\textit{V}}$. 

\subsubsection{Self-similarity Coherence Attention}
\label{sec:method_SCA}
RCA’s region-specific token influence neglects implicit physical coherence restrictions between neighboring objects, thus we propose SCA to explicitly integrate inherent physical coherence restriction into each pixel’s generation process. Specifically, SCA calculates the similarity between each pixel and its spatial adjacent feature within a defined local neighborhood region. As illustrated in Fig. \ref{fig:SFE}, it first reshapes the given image query to obtain $\textbf{\textit{Q}}_{r} \in \mathbb{R}^{C \times H \times W}$. Then, we apply zero-padding of size $(k-1)/2$ (assuming \textit{k} is odd) to $\textbf{\textit{Q}}_{r}$, resulting in the padded feature map $\Tilde{\textbf{\textit{Q}}}$ $\in \mathbb{R}^{C \times (H+k-1) \times (W+k-1)}$. This padding operation ensures that border pixels on the feature map are included in local self-similarity computation. After that, for any position $x_{ij}$ in the feature maps $\Tilde{\textbf{\textit{Q}}}$, where $x_{ij} \in \mathbb{R}^{C \times 1 \times 1}$, we construct a local neighbor region of spatial size $(k, k)$ centered at $\textbf{\textit{x}}_{ij}$, denoted as $\textbf{\textit{O}}_{ij}^{k\times k}$ $\in \mathbb{R}^{C \times k \times k}$. Next, we compute the dot product between $\textbf{\textit{x}}_{ij}$ and each neighboring position within the local region $\textbf{\textit{O}}_{ij}^{k\times k}$ and then sum the resulting vector across the channel dimension. This process yields the self-similarity vector $\overrightarrow{\textbf{\textit{ss}}}_{i,j} \in \mathbb{R}^{k^{2} \times 1}$. Then, all pixel's vectors constitute the self-similarity features map $\textbf{\textit{SS}} \in \mathbb{R}^{k^{2} \times H \times W}$. The process of generating the self-similarity feature map can be formulated as follows:
\begin{align}
	t&= (k - 1)/2,\\
	\overrightarrow{\textbf{\textit{ss}}}_{i,j} &= [\Tilde{\textbf{\textit{x}}}_{i-t,j-t}^{T} \cdot \textbf{\textit{x}}_{i,j}, \dots, \Tilde{\textbf{\textit{x}}}_{i+t,j+t}^{T} \cdot \textbf{\textit{x}}_{i,j}],\\
        \textbf{\textit{SS}} &= {\overrightarrow{\textbf{\textit{ss}}}_{1,1}, \dots, \overrightarrow{\textbf{\textit{ss}}}_{H,W}}.
\end{align}
Here, $\textbf{\textit{x}}_{i,j}$ and $\Tilde{\textbf{\textit{x}}}_{i,j}$ represent the spatial positions (\textit{i}, \textit{j}) within the feature map $\textbf{\textit{Q}}_{r}$ and the padded feature map $\Tilde{\textbf{\textit{Q}}}$, respectively. The hyperparameter \textit{k} denotes the length of the local neighbor region. $\overrightarrow{\textbf{\textit{ss}}}_{i,j}$ signifies the self-similarity vector at spatial position (\textit{i}, \textit{j}) within the feature map, and the self-similarity feature map \textbf{\textit{SS}}$\in \mathbb{R}^{k^{2} \times H \times W}$ is a aggregation of these self-similarity vectors $\overrightarrow{\textbf{\textit{ss}}}$.

Note that the local region size \textit{k} restricts the interaction scope of each pixel. This limitation results in the self-similarity feature map failing to encompass comprehensive contextual physical coherence information. To address this limitation, we employ convolution layers to expand the context and further capture intricate self-similarity patterns:
\begin{align}\label{M}
	\textbf{\textit{M}} = \textup{ReLU}(\textup{Conv}_{2}(\textup{ReLU}(\textup{Conv}_{1}(\textbf{\textit{SS}})))),
\end{align}
where \textbf{\textit{M}}$ \in\mathbb{R}^{N_{t} \times H \times W}$ is the generated self-similarity maps. Its empirical validations are presented in Section \ref{sec:SCA}.

Subsequently, we reshape and transpose the self-similarity maps and obtain \textbf{\textit{M}}$\in\mathbb{R}^{HW \times N_{t}}$. Finally, the output image feature $\textbf{\textit{I}}_{SCA} \in \mathbb{R}^{HW \times C_{x}}$ can be calculated by:       
\begin{align}\label{I}
	\textbf{\textit{I}}_{SCA} = \textup{Proj}(\textup{Softmax}(\textbf{\textit{M}})\textbf{\textit{V}}),
\end{align}
where Proj denotes the projection.

% The computation pipeline of SCA is illustrated in Algorithm \ref{alg:SCA}.
% \input{algorithm}

\subsubsection{Fusing Feature Maps}
To foster a dynamic interplay between RCA and SCA, SFE amalgamates the output maps $\textbf{\textit{I}}_{RCA}$ and $\textbf{\textit{I}}_{SCA}$ from them using a trainable parameter $\beta$, formulated as
\begin{align}
	\textbf{\textit{I}}_{SFE} = \textbf{\textit{I}}_{RCA} + \textup{tanh}(\beta) * \textbf{\textit{I}}_{SCA}.
\end{align}
This fusion empowers the model to leverage SCA's fine-grained understanding of potential local contextual coherence restrictions, thereby enhancing physical coherence generation. Simultaneously, it retains RCA's expertise in generating previously unseen objects.

% \subsection{Model Fine-tuning}
% The objective for fine-tuning is similar as that for pre-training (image denoising) \cite{rombach2022high}:
% \begin{equation}
%     	L(\theta) = \mathbb{E}_{\textbf{\textit{z}},l,\textbf{\textit{y}}_{c},\textbf{\textit{y}}_{t},\varepsilon \sim N(0,1), t} \bigg [ \parallel \varepsilon - \varepsilon_{\theta}(\textbf{\textit{z}}_{t}, t, l, c_{\phi}(\textbf{\textit{y}}_{c}), c_{\phi}(\textbf{\textit{y}}_{t}) \parallel_{2}^{2} \bigg] \nonumber
% \end{equation}
% Here, \textit{\textbf{z}} represents the latent code derived from the input image, \textit{l} signifies the input layout, $y_{c}$ denotes the input caption, $\textbf{\textit{y}}_{t}$ refers to the input text, $\varepsilon$ is a noise term, \textit{t} indicates the time step, $\varepsilon_{\theta}$ represents the denoising U-Net, $\textbf{\textit{z}}_{t}$ stands for the noisy version of \textit{\textbf{z}} at time \textit{t}, and $c_{\phi}$ symbolizes the text encoder. We exclusively fine-tune the denoising U-Net, while retaining the text encoder and the autoencoder of Stable Diffusion in a frozen state. Detailed training information can be found in the original paper of Stable Diffusion \cite{rombach2022high}.

%% file: figs/performance.tex
% Use figure* for multi-column figure
\begin{figure*}[ht]
    \centering
    \includegraphics[width=0.8\linewidth]{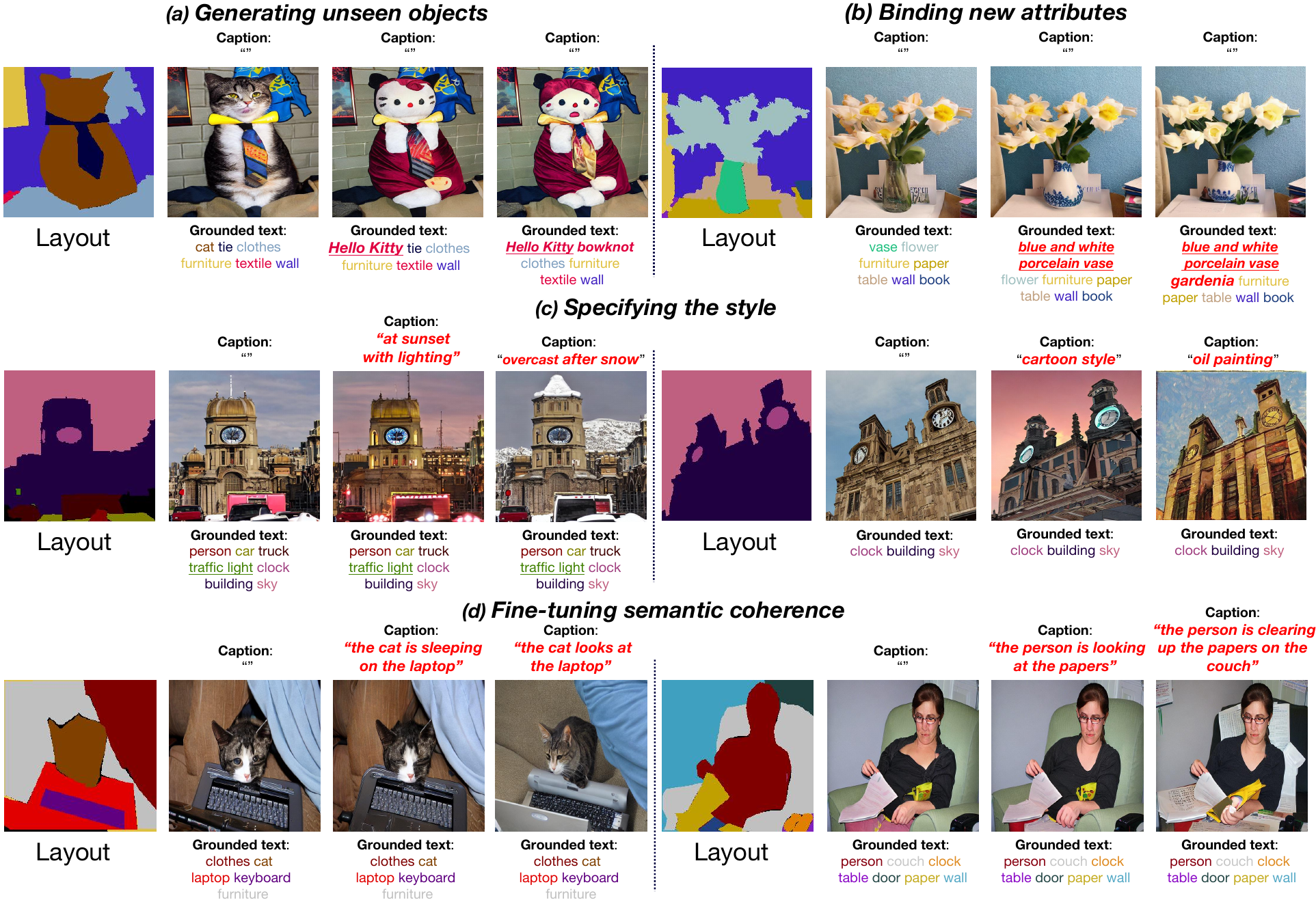}
    \caption{Exemplar results and demonstrated capabilities of our EOCNet. Zooming in for better visibility and detail.}
    \label{fig:performance}
    \vspace{-0.3cm}
\end{figure*}

%% file: 5_exp.tex
\section{Experiments}
\label{sec:experiments}

\input{figs/coco_ade20k}

\subsection{Implementation details}
\noindent\textbf{\textit{Datasets.}} We employ two datasets, COCO-Stuff \cite{caesar2018coco} and ADE20K \cite{zhou2017scene}. COCO-Stuff comprises 118,287 training and 5,000 validation images, annotated with 182 semantic classes. ADE20K contains 20,210 training and 2,000 validation images, encompassing 150 semantic classes. 

% \noindent\textbf{\textit{Training configures.}} We adopt the origin training set of Stable Diffusion \cite{rombach2022high} unless specific noted and fine-tune our model based on the \textit{stable-diffusion-v1.4}. The base learning rate is $10^{-6}$. We use the BLIP-2 model \cite{li2023blip} \textit{blip2-opt-6.7b-coco} to generate captions for all images in COCO-Stuff and train our model on COCO-Stuff/ADE20K takes approximately 4/1 days on 4 NVIDIA A100 GPUs. We use 50 PLMS \cite{liu2022pseudo} steps with a scale of 2 for classifier-free guidance \cite{ho2022classifier}.

% \noindent\textbf{\textit{Evaluation metrics.}} We evaluate segmentation accuracy using mean Intersection over-Union (mIoU). To evaluate the overall fidelity, we employ Fréchet Inception Distance (FID) \cite{heusel2017gans}and CLIP-I to measure the distribution distance between synthesized and real images, and the average pairwise cosine similarity between their CLIP embeddings.  To assess semantic coherence, we employ CLIP-T to compute the average cosine similarity between prompt and image CLIP embeddings. We select 100 images from the COCO-Stuff validation set, manually annotate physical coherence areas using bounding boxes, and then use DINO to evaluate physical coherence by calculating the average pairwise cosine similarity between the ViT-S/16 DINO embeddings of the generated and real physical coherence areas.
% For diversity comparison. We assess diversity using Diversity Score (DS) akin to \cite{oasis}. 

\noindent\textbf{\textit{Baselines.}} We compare our method against the state-of-the-art LIS baselines including Pix2PixHD \cite{P2PHD}, SPADE \cite{SPADE}, CC-FPSE \cite{CC-FPSE}, OASIS \cite{oasis}, SC-GAN \cite{SC-GAN}, PITI \cite{PITI}, ControlNet \cite{controlnet} and FreestyleNet \cite{freestyleNet}. We employ pre-trained models released from their source code and evaluate them according to the settings described in their original papers.

\input{tables/compare}

\subsection{Comparison}
\label{sec:comparison}
\textbf{\textit{Quantitative results.}} 
As illustrated in Table \ref{tab:compare} and Table \ref{tab:compare_clip}, our method surpasses existing ones in all metrics. 
% overall image quality (FID and CLIP-I), diversity (DS), semantic coherence (CLIP-T), and physical coherence (DINO). 
% This demonstrates the superior effectiveness of our approach.
These results collectively highlight the effectiveness of our approach in enhancing object coherence and underscore the remarkable visual quality and diversity of the images generated through our approach. Notably, the recent FreestyleNet also showcases commendable generation quality. However, its reliance solely on RCA to constrain text token influence within designated regions leads to the omission of potential physically coherent relationships between objects in the layout. As a consequence, its realism and naturalness in terms of physical coherence generation are compromised. Our model yields comparatively lower mIoU scores in contrast to certain other LIS methods. This observation is in line with the argument presented in \cite{freestyleNet}, where it is asserted that mIoU might not be an entirely suitable metric for evaluating models involving RCA. This limitation arises from the potential generation of more generalized semantics driven by pre-trained knowledge (e.g., the building in the 3rd row of Fig. \ref{fig:coco_ade20k}), consequently leading to misclassifications of class labels by the segmentation model employed in calculating mIoU.

\noindent\textbf{\textit{Qualitative results.}} Our approach consistently produces images with heightened sharpness and intricate details. Taking the examples in Fig. \ref{fig:coco_ade20k} (first row), our method notably achieves superior physical coherence between the person and the baseball bat. In contrast, certain GAN-based baselines, such as SPADE, OASIS, and SC-GAN, encounter challenges in synthesizing the person effectively. This might be attributed to inherent limitations in the generative capacity of GANs.
While other diffusion-based methods like ControlNet \cite{controlnet} and FreestyleNet \cite{freestyleNet} show improved generation of the person, they still grapple with the physical coherence problem due to their inability to capture potential physical coherence restrictions. This observation underscores the effectiveness of our approach in improving physical coherence generation.

\input{tables/compare_clip}

% This dual effect underscores SCA's role in bolstering both physical coherence and intricate texture synthesis. 

%% file: figs/coco_ade20k.tex
\begin{figure*}[th!]
    \centering
    \includegraphics[width=0.88\linewidth]{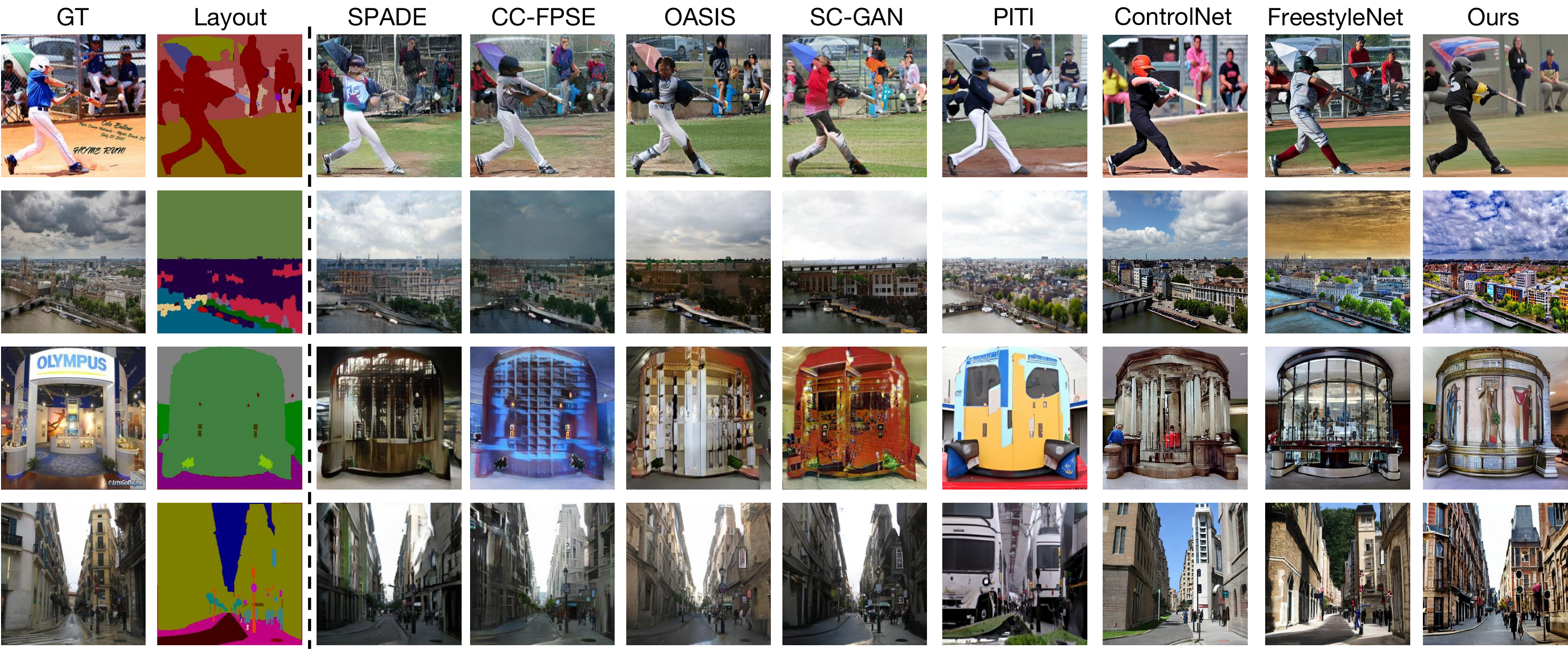}
    \caption{Qualitative comparative results on COCO-Stuff (upper 2 rows) and ADE20K (lower 2 rows).}
    \label{fig:coco_ade20k}
    \vspace{-0.2cm}
\end{figure*}

%% file: tables/compare.tex
\begin{table}[t]
\setlength\tabcolsep{9pt}
\centering
% \resizebox{\columnwidth}{!}{%
\tiny
\begin{tabular}{cccc|ccc}

		\bottomrule
		\multirow{2}{*}{\textbf{Methods}} & \multicolumn{3}{c|}{\textit{\textbf{COCO-stuff}}} & \multicolumn{3}{c}{\textit{\textbf{ADE20K}}} \\
		& FID  $\downarrow$       & mIoU $\uparrow$  &DS $\uparrow$         & FID $\downarrow$ & mIoU $\uparrow$ &DS $\uparrow$   \\
		\bottomrule
		Pix2PixHD        & 110.7             & 15.2   &-      & 82.3 & 20.8     &-   \\
		SPADE              & 23.2             & 36.8     &-    & 33.6 & 38.2 &-   \\
		CC-FPSE            & 20.0             & 41.5  & 0.079     & 32.3 & 42.5  &0.118   \\
		OASIS              & 16.9             & \textbf{43.9}  &0.328       & 27.8  & \textbf{48.1} &0.282  \\
	       SC-GAN         & 17.7             & 41.8   &-      & 28.4 & 44.5  &-   \\
            PITI         & 15.6             & 35.8   &0.533      & 27.3 & 30.6    & 0.492\\
            ControlNet          & 15.4             & 42.9   &0.602      & 26.8 & 39.9    & 0.541\\
            FreestyleNet         & 15.1             & 41.5  &0.591       & 25.7 & 42.0  &0.587   \\
            \textbf{EOCNet}(Ours)          & \textbf{14.2}             & 43.0    &\textbf{0.624}     & \textbf{24.6} & 43.3 &\textbf{0.619}    \\
		\bottomrule
	\end{tabular} \\

% }
\caption{Quantitative comparison results. 
% '-' means the data is not available. 
% Pix2PixHD,SPADE and SC-GAN do not support diverse generations.
}
\vspace{-0.3cm}
\label{tab:compare}
\end{table}

%% file: tables/compare_clip.tex
\begin{table}[t]
\setlength\tabcolsep{19pt}
\centering
% \resizebox{\columnwidth}{!}{%
\tiny
\begin{tabular}{cccc}

		\bottomrule
		\textbf{Methods} & CLIP-I  $\uparrow$       & CLIP-T $\uparrow$  &DINO $\uparrow$           \\
		\bottomrule
		Pix2PixHD            & 0.533             & 0.11   &0.374         \\
		SPADE              & 0.698             & 0.14     &0.436       \\
		CC-FPSE              & 0.731             & 0.17  & 0.488        \\
		OASIS           & 0.766             & 0.18  &0.503         \\
	       SC-GAN         & 0.753             & 0.20   &0.511         \\
            PITI         & 0.796            &   0.23 &0.576      \\
            ControlNet         & 0.803             & 0.28   &0.628      \\
            FreestyleNet          & 0.812             & 0.26  &0.634          \\
            \textbf{EOCNet}(Ours)          & \textbf{0.831}             & \textbf{0.32}    &\textbf{0.670}         \\
		\bottomrule
	\end{tabular} \\

% }
\caption{Quantitative comparison results.}
\vspace{-0.3cm}
\label{tab:compare_clip}
\end{table}

%% file: 7_conclusion.tex
\section{Conclusion}
\label{sec:conclusion}
This work addresses the object coherence challenge within the LIS task.
% including semantic coherence and physical coherence. 
We propose a novel diffusion model with GSF and SFE. For semantic coherence, we introduce caption as input and develop GSF to fuse the supervision from the layout restriction and semantic coherence requirement and exploit it to guide the image synthesis process. To improve the physical coherence generation, we propose SFE that leverages the effective synergy between RCA and SCA. 
% It employs SCA to explicitly integrate local contextual physical coherence restriction into each pixel’s generation process while retaining RCA’s superiority in generating previously unseen objects. 
Comprehensive experiments showcase our superior image quality and enhanced controllability.

\section{Acknowledgments}
This work was supported in part by the National Natural Science Foundation of China under Grant 62406285 and the China Postdoctoral Science Foundation under Grant 2024M752863.

%% file: 8_appendix.tex
\appendix
\label{sec:appendix}
\subsection{Algorithm}
The computation pipeline of SCA is illustrated in Algo. \ref{alg:SCA}.
\input{algorithm}

\subsection{Implementation details}
\noindent\textbf{\textit{Training configures.}} We adopt the origin training set of Stable Diffusion \cite{rombach2022high} unless specific noted and fine-tune our model based on the \textit{stable-diffusion-v1.4}. The base learning rate is $10^{-6}$. We use the BLIP-2 model \cite{li2023blip} \textit{blip2-opt-6.7b-coco} to generate captions for all images in COCO-Stuff and train our model on COCO-Stuff/ADE20K takes approximately 4/1 days on 4 NVIDIA A100 GPUs. We use 50 PLMS \cite{liu2022pseudo} steps with a scale of 2 for classifier-free guidance \cite{ho2022classifier}.

\noindent\textbf{\textit{Evaluation metrics.}} We evaluate segmentation accuracy using mean Intersection over-Union (mIoU). To evaluate the overall fidelity, we employ Fréchet Inception Distance (FID) \cite{heusel2017gans}and CLIP-I to measure the distribution distance between synthesized and real images, and the average pairwise cosine similarity between their CLIP embeddings.  To assess semantic coherence, we employ CLIP-T to compute the average cosine similarity between prompt and image CLIP embeddings. We select 100 images from the COCO-Stuff validation set, manually annotate physical coherence areas using bounding boxes, and then use DINO to evaluate physical coherence by calculating the average pairwise cosine similarity between the ViT-S/16 DINO embeddings of the generated and real physical coherence areas.
For diversity comparison. We assess diversity using Diversity Score (DS) akin to \cite{oasis}. 

\subsection{Qualitative Evaluation}
\noindent\textbf{\textit{Superiority compared with SOTA.}} Current SOTA LIS diffusion models \cite{freestyleNet} utilize semantic masks to determine object positioning and employ textual inputs to control object classes. Our model inherits these functionalities while introducing novel enhancements to achieve greater controllability and higher generation quality. Specifically, by introducing the caption as input and utilizing our GSF to integrate semantic scene or overall style requirements into the generation process, our model gains the power to finely tune semantic coherence in the resulting image and control its style. On the other hand, we also propose SFE that leverages SCA’s fine-grained understanding of local contextual coherence relations to improve physical coherence generation.

\noindent\textbf{\textit{Qualitative Results.}} Through encapsulating objects' classes within the input text and specifying semantic coherence requirements in the caption, our model effectively synthesizes compelling images. We highlight four distinct capabilities of our proposed model in Fig. \ref{fig:performance}, placing particular emphasis on the caption's role in style specification (2nd row) and semantic coherence tuning (3rd row). The synergistic results of all capabilities are demonstrated in Fig. \ref{fig:display}. The effectiveness of our model in addressing object coherence challenges is shown in Fig. \ref{fig:problem}. These results demonstrate that our model faithfully generates images portraying the novel objects and semantic scenes described in the input text and caption respectively. It's important to note again that our caption can not only specify the image style but also adjust the semantic coherence between objects. For instance, the 3rd row of Fig. \ref{fig:performance}, demonstrates how the cat interacts with the laptop as dictated by the caption.

\subsection{Diversity evaluation}\label{diversity}
Our model naturally facilitates diverse image generation from a single layout using different texts and captions (refer to Fig. 1 and Fig. 5 in the main paper). We assess diversity akin to OASIS \cite{oasis}. Specifically, the Diversity Score (DS) is computed between images generated from identical layouts (and the same text for our model) with randomly sampled noise. Evaluation results are detailed in Tab. 1 in the main paper, where our model achieves the highest DS compared to all other methods. Here, we provide some visual examples of our model in Fig. \ref{fig:diversity}.

\subsection{User Study}
We conduct a user study to compare five methods: EOCNet, FreestyleNet \cite{freestyleNet}, ControlNet \cite{controlnet}, PITI \cite{PITI}, and SC-GAN \cite{SC-GAN}. We invite 20 users and task them with completing 40 ranking questions. For each question, the participants are presented with five images generated using different methods. They are requested to rank five images from 1 to 5 (lower is worse) based on "the overall quality of displayed images" and "the fidelity of physical coherence". The average rankings are shown in Tab. \ref{tab:user_study}.

% \subsection{Dual Effectiveness of SCA}
% \label{sec:SCA}

\subsection{Ablation Studies}
\noindent\textbf{SCA}: 
\noindent\ \textit{Effectiveness on physical coherence generation.} Upon meticulous examination of the visualized self-similarity maps of SCA in Fig. \ref{fig:SCA}, a discernible pattern emerges: SCA prominently emphasizes values, particularly in object coherence, such as the joint region of the hand and racket (first sample in Fig. \ref{fig:SCA}). This observation underscores SCA's effectiveness in enhancing physical coherence within generated images. 
\input{figs/SCA_effective}

\noindent\textit{Effectiveness on complex texture generation.} Furthermore, these visualizations also highlight SCA's impact on intricate texture generation, evident in emphasized values also within regions with complex texture like facial details, seen in the girl's face region (second sample in Fig. \ref{fig:SCA}).

We attribute this to the intricate diversity within challenging-to-generate objects like a person's face and fingers. Achieving effective synthesis in these regions demands models to explore more subtle correlations between pixels. However, existing methods only utilize cross-attention to capture complex correlations, which might be insufficient, resulting in their failure to generate subtle texture variations.
% resulting in the incomplete or unsatisfactory generation of complex objects.

However, our SNF exhibits exceptional sensitivity and adaptability to the diversity and complexity within these regions. This enables our approach to proficiently capture inter-class coherence correlations and intra-class intricate textures, which significantly promotes complex texture generation. We additionally offer qualitative comparisons on complex textures generation with FreestyleNet in Fig. \ref{fig:complex_texture}.

To quantitatively evaluate the impact of SCA, we conduct an ablation study, with the results shown in Table \ref{tab:ablation}. Notably, with the addition of SCA, the model's performance significantly improved on the DINO metric, demonstrating its crucial role in achieving better physical coherence. Furthermore, the results also reveal that SCA contributes positively to other metrics such as FID and CLIP-I, indicating that the enhancement in physical coherence has a beneficial ripple effect on overall image quality. 
The qualitative ablation results are provided in Fig. \ref{fig:SCA_effectiveness}.
\input{figs/complex_texture}

\noindent\textbf{GSF}: GSF fuses supervision information from the layout restriction and caption requirement, and infuses the amalgamated supervision features into the noise image to steer the image synthesis effectively, thereby enhancing semantic coherence control over image generation. 

To validate its effectiveness quantitatively, we conduct ablation experiments. As shown in Table \ref{tab:ablation}, upon adding this module, the model's performance is improved across all metrics. For example, there is a notable increase of 0.11 and 0.04 in CLIP-I and CLIP-T, respectively, indicating a significant improvement in image quality and semantic coherence. This result strongly underscores the pivotal role of the GSF module in image synthesis. 

We conduct qualitative ablation experiments by directly removing GSF to evaluate the effectiveness of GSF. Notably, in its absence, we observed the stylistic inconsistency problem within generated images as shown in Fig. \ref{fig:GSF_effective}. Certain objects exhibited styles that deviated from the overall scene, leading to unsatisfied image generation. This observation demonstrated GSF's additional effectiveness in maintaining the style consistency of generated images.

Further, we experiment with a naive approach to control semantic coherence in generated images. This involved concatenating the caption and grounded text, allowing the semantic information to interact with noisy images within cross-attention layers. The qualitative results are depicted in Fig. \ref{fig:ablation_GSF_CA}. It is evident that this method is ineffective as it addresses the highly relevant layout restriction and semantic coherence requirement separately while the relative positions of objects in the layout would inevitably influence their semantic coherence interactions.

These experimental results undoubtedly demonstrated the indispensable role of GSF in elevating both the controllability and synthesis quality, aligning with the primary goals of our research.
% The qualitative ablation results are provided in the supplementary.
\input{tables/ablation_experiment}

\subsection{More visualized self-similarity maps} 
We provide more visualized self-similarity maps in Fig. \ref{fig:SCA_supplementary}.

% \subsection{Effectiveness of SCA}\label{effectiveness_SCA}
% The qualitative ablation results are provided in Fig. \ref{fig:SCA_effectiveness}.

\input{tables/user_study}

% \subsection{Effectiveness of GSF}\label{effectiveness_GSF}
% We conduct qualitative ablation experiments by directly removing GSF to evaluate the effectiveness of GSF. Notably, in its absence, we observed the stylistic inconsistency problem within generated images as shown in Fig. \ref{fig:GSF_effective}. Certain objects exhibited styles that deviated from the overall scene, leading to unsatisfied image generation. This observation demonstrated GSF's additional effectiveness in maintaining the style consistency of generated images.

% Further, we experiment with a naive approach to control semantic coherence in generated images. This involved concatenating the caption and grounded text, allowing the semantic information to interact with noisy images within cross-attention layers. The qualitative results are depicted in Fig. \ref{fig:ablation_GSF_CA}. It is evident that this method is ineffective as it addresses the highly relevant layout restriction and semantic coherence requirement separately while the relative positions of objects in the layout would inevitably influence their semantic coherence interactions.

% These experimental results undoubtedly demonstrated the indispensable role of GSF in elevating both the controllability and synthesis quality, aligning with the primary goals of our research.

\subsection{More qualitative comparisons with LIS baselines} \label{more_comparision}
We provide more qualitative comparative results against the LIS baselines on COCO-stuff and ADE20K in Fig. \ref{fig:more_coco} and Fig. \ref{fig:more_ade20k} respectively.

\input{6_limitation}
\input{figs/failure}

\subsection{Societal impact} \label{societal impact}
Our work has a multifaceted societal impact, offering opportunities for creative expression and responsible application. Our method facilitates diverse image generation via layout, text, and caption manipulation, which unleashes users' power to enhance their creativity for more personalized image creation. This innovation opens new horizons in content creation and artistic exploration, with potentially transformative effects on industries like entertainment, advertising, and design that seek tailored and contextually rich visuals.

\input{figs/supply_physical_coherence}
\input{figs/diversity}
\input{figs/SCA_supplementary}
\input{figs/GSF_effective}

\input{figs/ablation_GSF_CA}

\input{figs/more_coco}
\input{figs/more_ade20k}

%% file: algorithm.tex
% \subsection{Algorithm}\label{SCA}
% The computation pipeline of SCA is illustrated in \cref{alg:SCA}.
\begin{algorithm}[!ht]
    \caption{SCA}
    \label{alg:SCA}
    \renewcommand{\algorithmicrequire}{\textbf{Input:}}
    \renewcommand{\algorithmicensure}{\textbf{Output:}}
    \begin{algorithmic}
        \REQUIRE Input Query $\textbf{\textit{Q}}$, Value $\textbf{\textit{V}}$, local context size $k$ %%input
        \ENSURE Output image feature $\textbf{\textit{I}}_{SCA}$    %%output
        
        \STATE  Reshape $\textbf{\textit{Q}}$ to match $\mathbb{R}^{C \times H \times W}$
        \STATE  Get padded feature $\Tilde{\textbf{\textit{Q}}} \in \mathbb{R}^{C \times (H+k-1) \times (W+k-1)}$ by adding zero-padding of size ($k$-1)/2 to $\textbf{\textit{Q}}$ 
        \STATE  Initialize an feature map $\textbf{\textit{SS}} \in \mathbb{R}^{k^2 \times H \times W}$
        
        \FOR{each $i$ in {0,1,...,$H$-1}} 
            \FOR{each $j$ in {0,1,...,$W$-1}}
                \STATE $\textbf{\textit{O}} \in \mathbb{R}^{C, k^{2}} \gets$ the flatten features of local (\textit{k} $\times$ \textit{k}) region centered at the spatial position$(i,j)$ in $\Tilde{\textbf{\textit{Q}}}$
                \FOR{each $t$ in [0,1,...,$k^{2}$]}
                    \STATE $\textbf{\textit{SS}}_{t, i, j} \gets$ $\sum_{c=0}^C$$\textbf{\textit{Q}}_{c,i,j}*\textbf{\textit{O}}_{c,t}$. 
                \ENDFOR
            \ENDFOR
        \ENDFOR
        \STATE Get the self-similarity maps $\textbf{\textit{M}}$ by Eq. 9.
        \STATE Get the output image feature maps $\textbf{\textit{I}}_{SCA}$ by Eq. 10.
    \end{algorithmic}
\end{algorithm}

%% file: figs/SCA_effective.tex
\begin{figure}[t]
    \centering
    \includegraphics[width=0.75\linewidth]{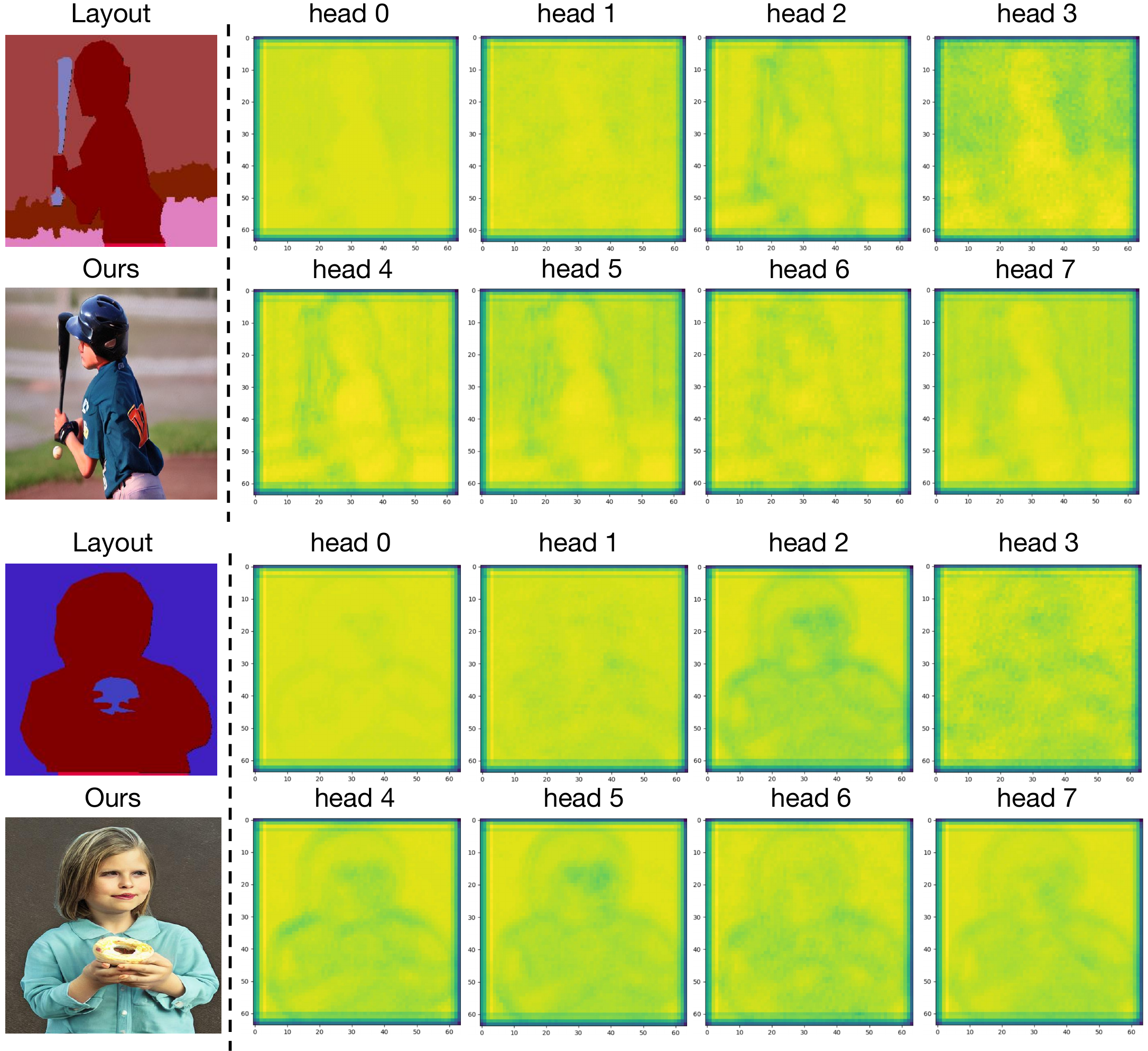}
    \caption{Effectiveness of the proposed SCA. We present visualized average self-similarity maps for each head. 
    % Notably, these heatmaps emphasize significant values in the coherence of objects and regions with intricate textures. We omit the grounded text and captions for simplicity.
    }
    \label{fig:SCA}
    \vspace{-0.2cm}
\end{figure}

%% file: figs/complex_texture.tex
\begin{figure}[t]
    \centering
    \includegraphics[width=0.9\linewidth]{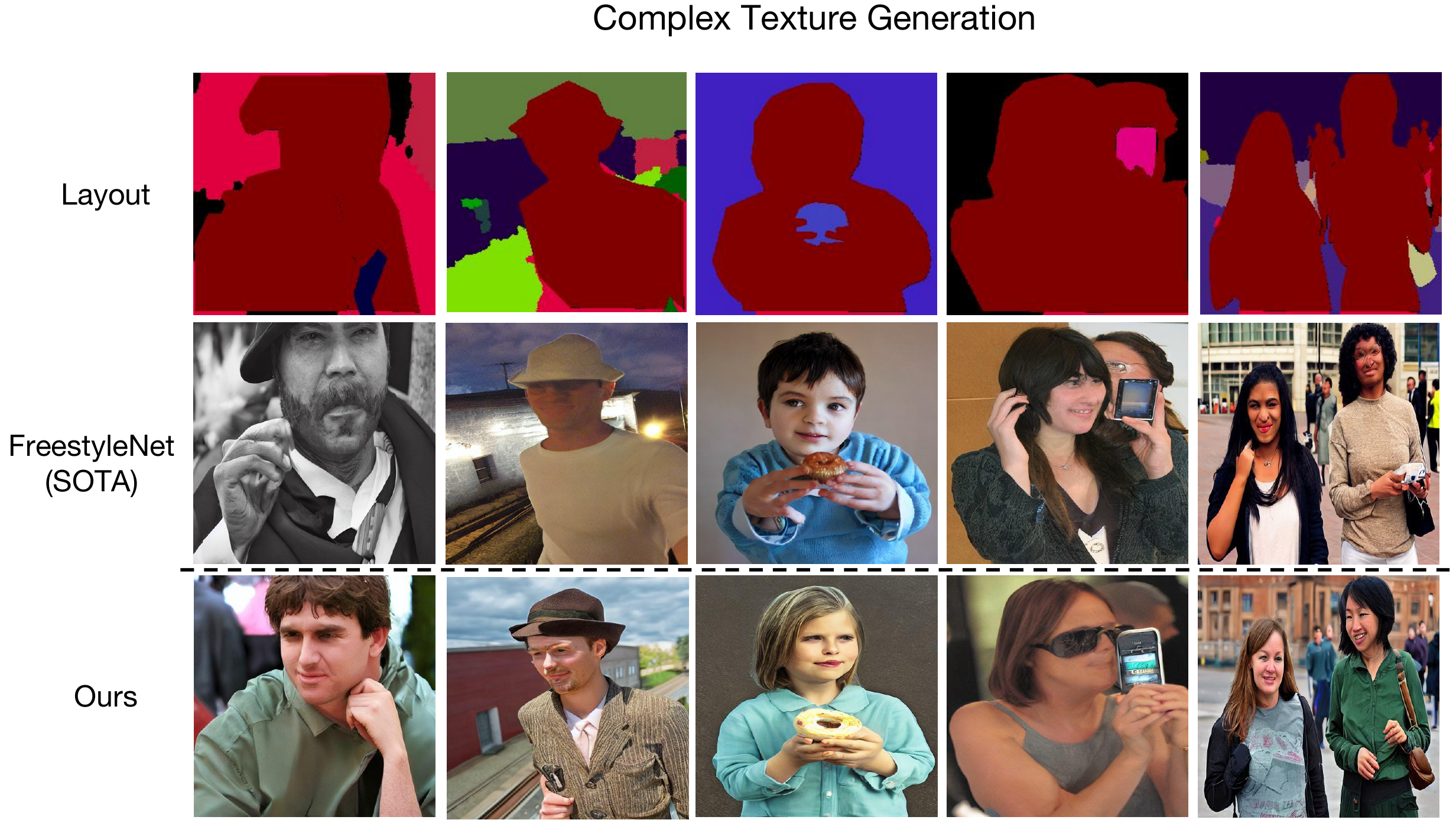}
    \caption{Qualitative comparative results on complex texture generation against the FreestyleNet. We omit the grounded text for simplicity.}
    \label{fig:complex_texture}

\end{figure}

%% file: tables/ablation_experiment.tex
\begin{table}[t]
\setlength\tabcolsep{8pt}
\centering
% \resizebox{\columnwidth}{!}{%
% \scriptsize
\begin{tabular}{ccccc}

		\bottomrule
		\textbf{Methods} & FID $\downarrow$& CLIP-I  $\uparrow$       & CLIP-T $\uparrow$  &DINO $\uparrow$           \\
		\bottomrule
		Baseline+RCA          &15.1  & 0.812             & 0.26   &0.634         \\
		+GSF  &   14.7          & 0.823             & 0.31     &0.641       \\
            +SCA(\textbf{Ours})       &\textbf{14.2}   & \textbf{0.831}             & \textbf{0.32}    &\textbf{0.670}         \\
		\bottomrule
	\end{tabular} \\

% }
\caption{Ablation results on SCA and GSF.}

\label{tab:ablation}
\end{table}

%% file: tables/user_study.tex
\begin{table}[t]
\setlength\tabcolsep{15pt}
\centering
% \resizebox{\columnwidth}{!}{%

\begin{tabular}{ccc}

		\bottomrule
		\textbf{Methods} & Overall Quality $\uparrow$       & PC Fidelity $\uparrow$            \\
		\bottomrule
	       SC-GAN          & 1.557             &  2.195       \\
            PITI        & 2.687            &   2.829      \\
            ControlNet         & 3.173             & 3.568        \\
            FreestyleNet          & 3.462             & 3.083            \\
            \textbf{EOCNet}(Ours)          & \textbf{4.121}             & \textbf{4.325}            \\
		\bottomrule
	\end{tabular} \\

% }
\caption{User study of Overall Quality and Physical Coherence (PC) Fidelity: higher score, better ranking. }
% \vspace{-0.7cm}
\label{tab:user_study}
\end{table}

%% file: 6_limitation.tex
\subsection{Limitation}
Several failure cases of our method are illustrated in Fig. \ref{fig:failure}. Our approach showcases enhanced capabilities in most object coherence and complex texture generation. However, we fail to generate certain scenarios well, such as generating extremely intricate textures resembling hands (Fig. \ref{fig:failure} (left)). Moreover, despite the integration of caption-guided semantic coherence control in our method, our model struggles to achieve semantic alignment when conflicts arise between the semantic coherence requirement specified in the caption and the underlying layout (Fig. \ref{fig:failure} (right)). 
Besides, due to the inherent limitations of the stable diffusion model, the generated results will inevitably exhibit some artifacts.
These limitations underscore the need for further exploration.

%% file: figs/failure.tex
\begin{figure}[ht]
    \centering
    \includegraphics[width=0.9\linewidth]{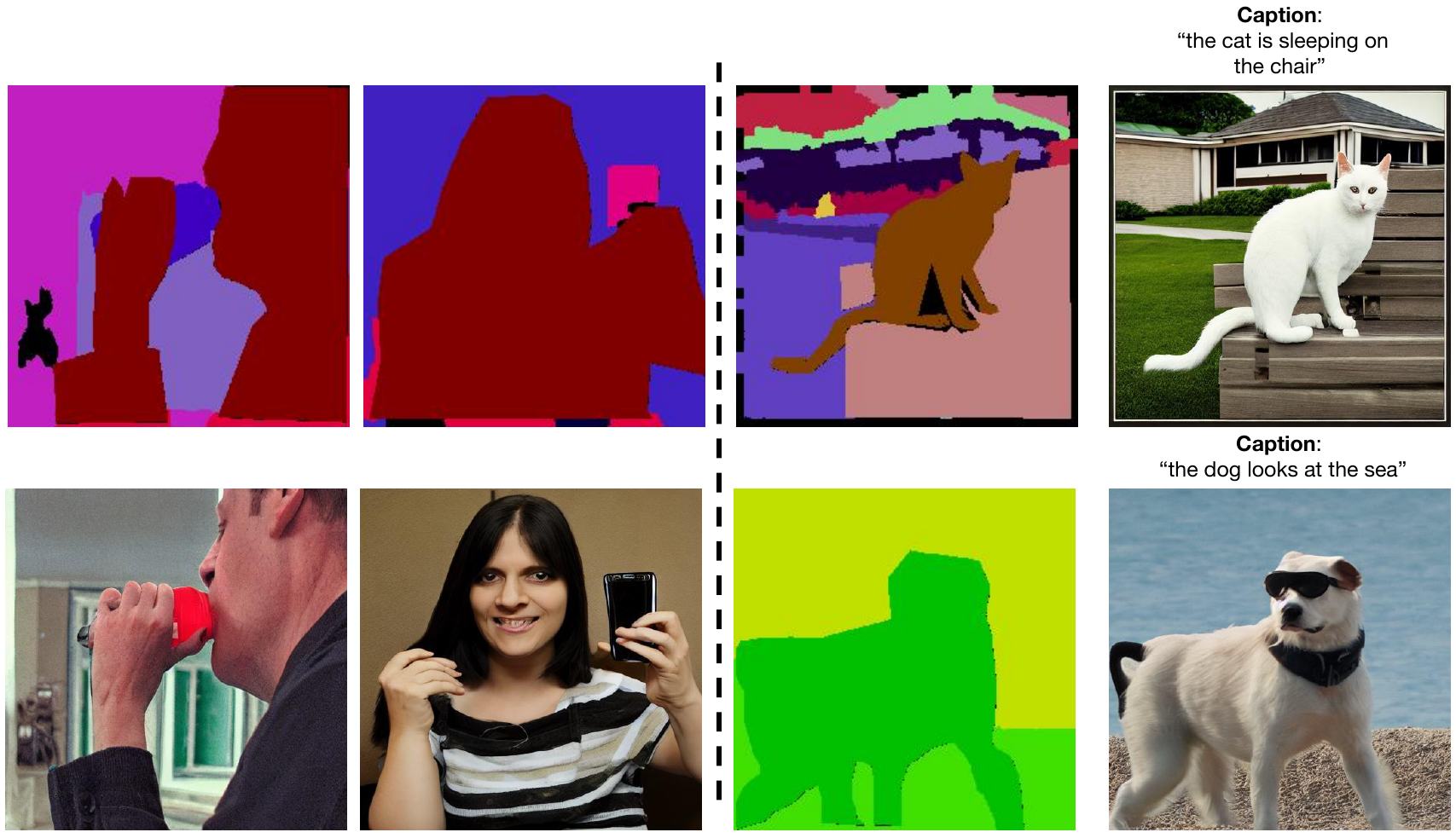}
    \caption{Failure cases.}
    \label{fig:failure}
    \vspace{-0.3cm}
\end{figure}

%% file: figs/supply_physical_coherence.tex
\begin{figure*}[ht]
    \centering
    \includegraphics[width=0.62\linewidth]{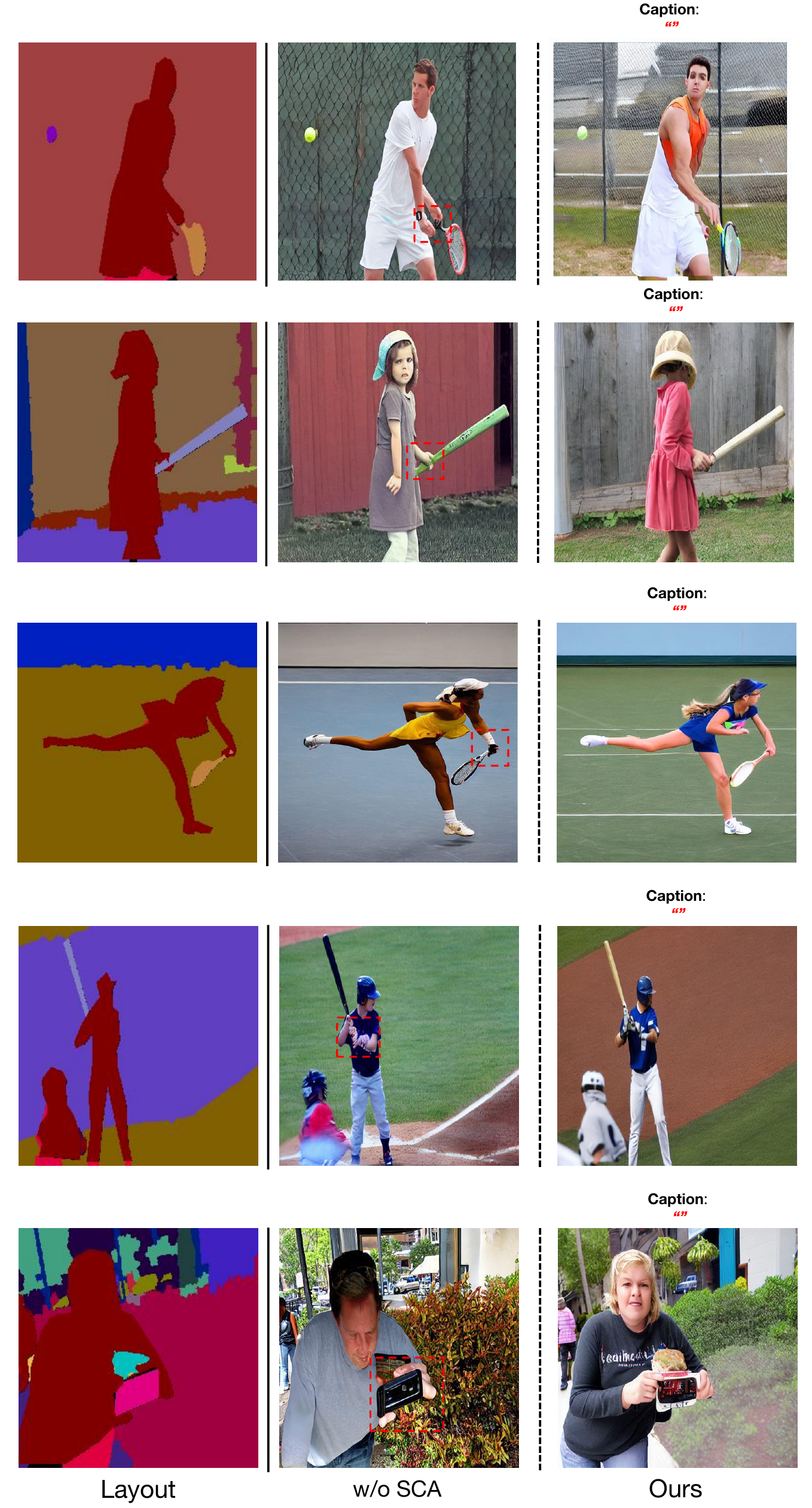}
    \caption{Effectiveness of the proposed SCA. We omit the grounded text for simplicity.}
    \label{fig:SCA_effectiveness}
\end{figure*}

%% file: figs/diversity.tex
\begin{figure*}[ht]
    \centering
    \includegraphics[width=\linewidth]{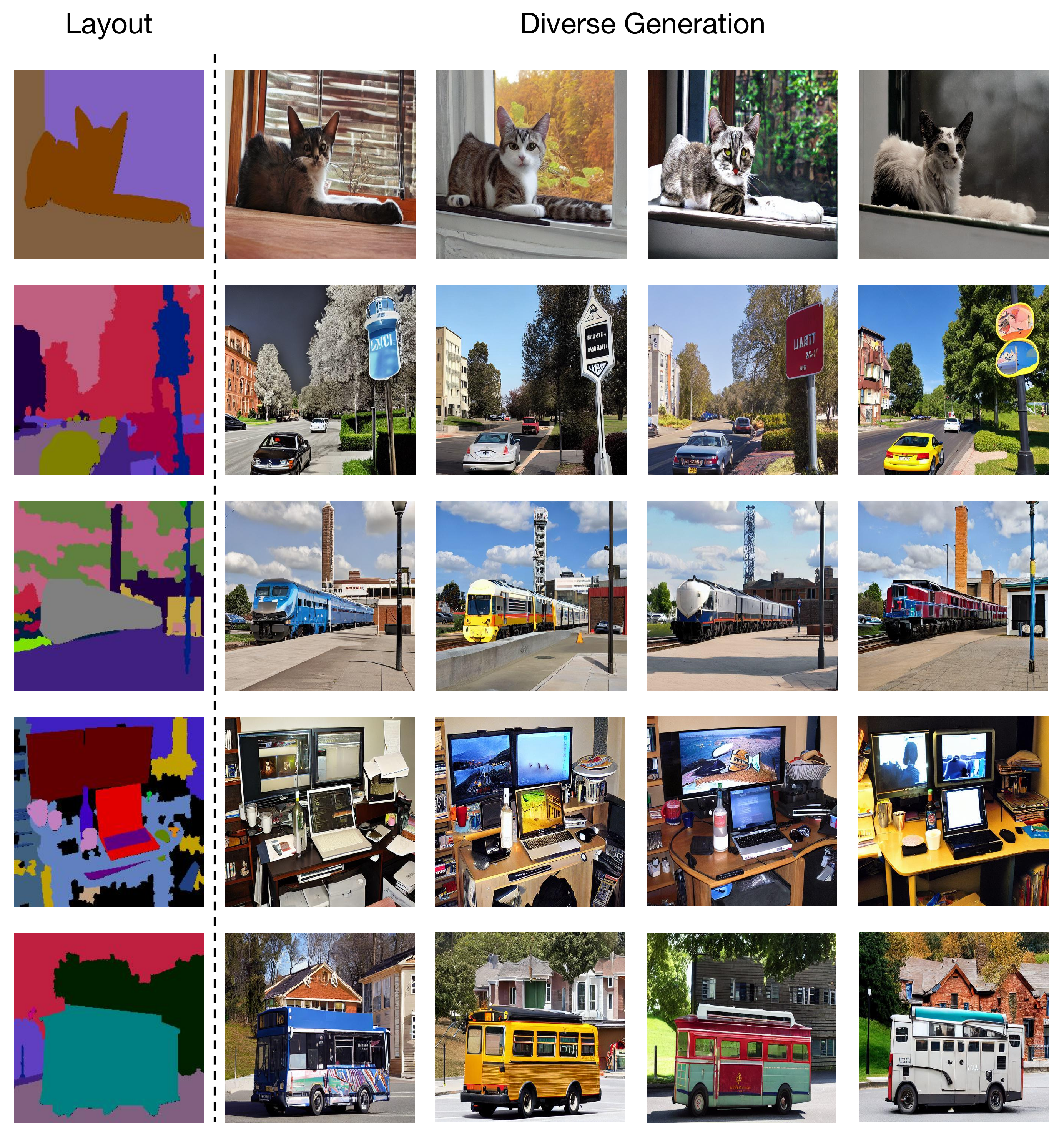}
    \caption{Diverse generation results of our proposed model. The captions in these cases are set to empty and we omit the grounded text for simplicity.}
    \label{fig:diversity}
\end{figure*}

%% file: figs/SCA_supplementary.tex
\begin{figure*}[ht]
    \centering
    \includegraphics[width=0.9\linewidth]{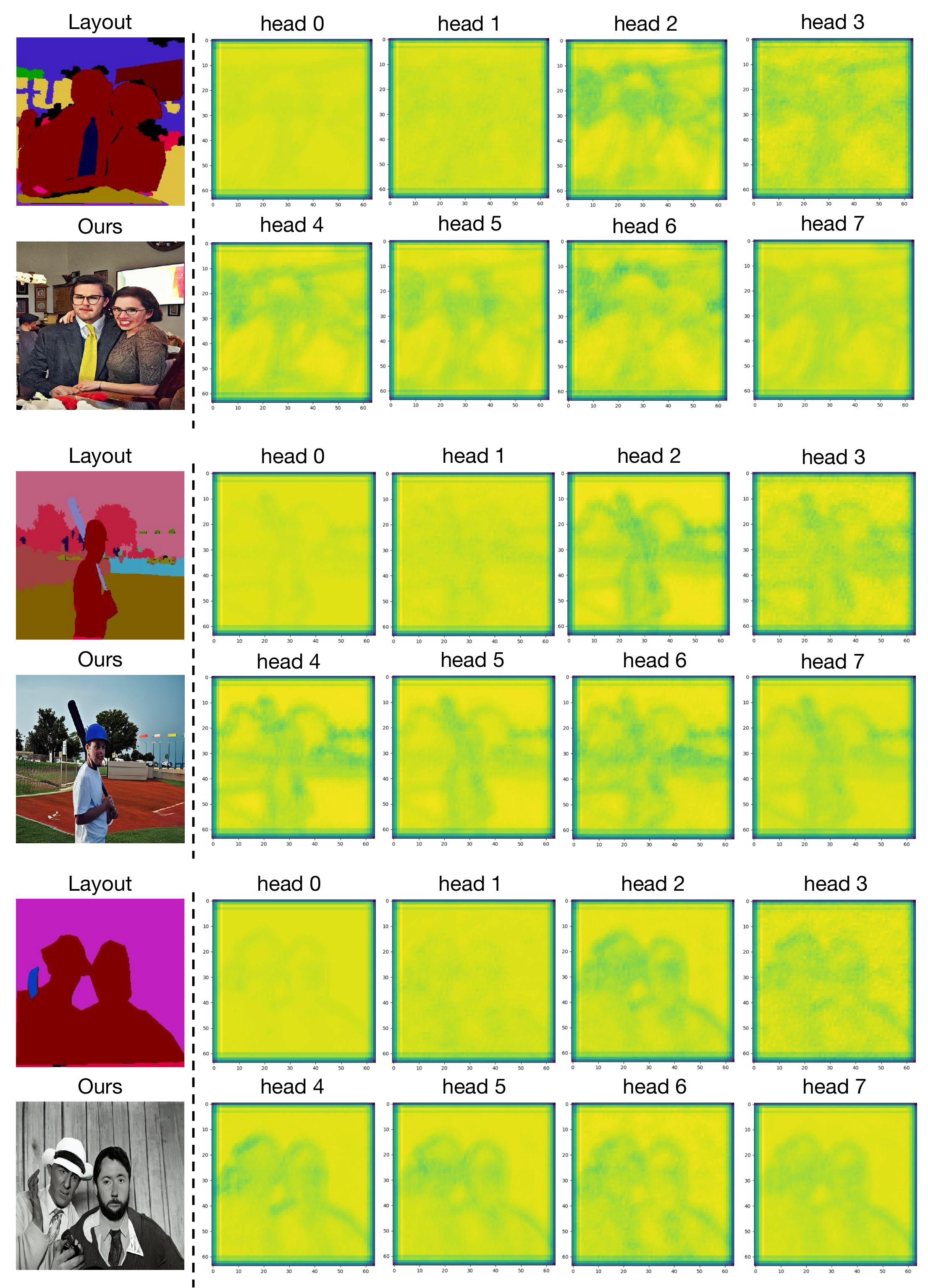}
    \caption{We present more visualized self-similarity maps. The captions in these cases are set to empty and we omit the grounded text for simplicity.}
    \label{fig:SCA_supplementary}
\end{figure*}

%% file: figs/GSF_effective.tex
\begin{figure*}[ht]
    \centering
    \includegraphics[width=0.66\linewidth]{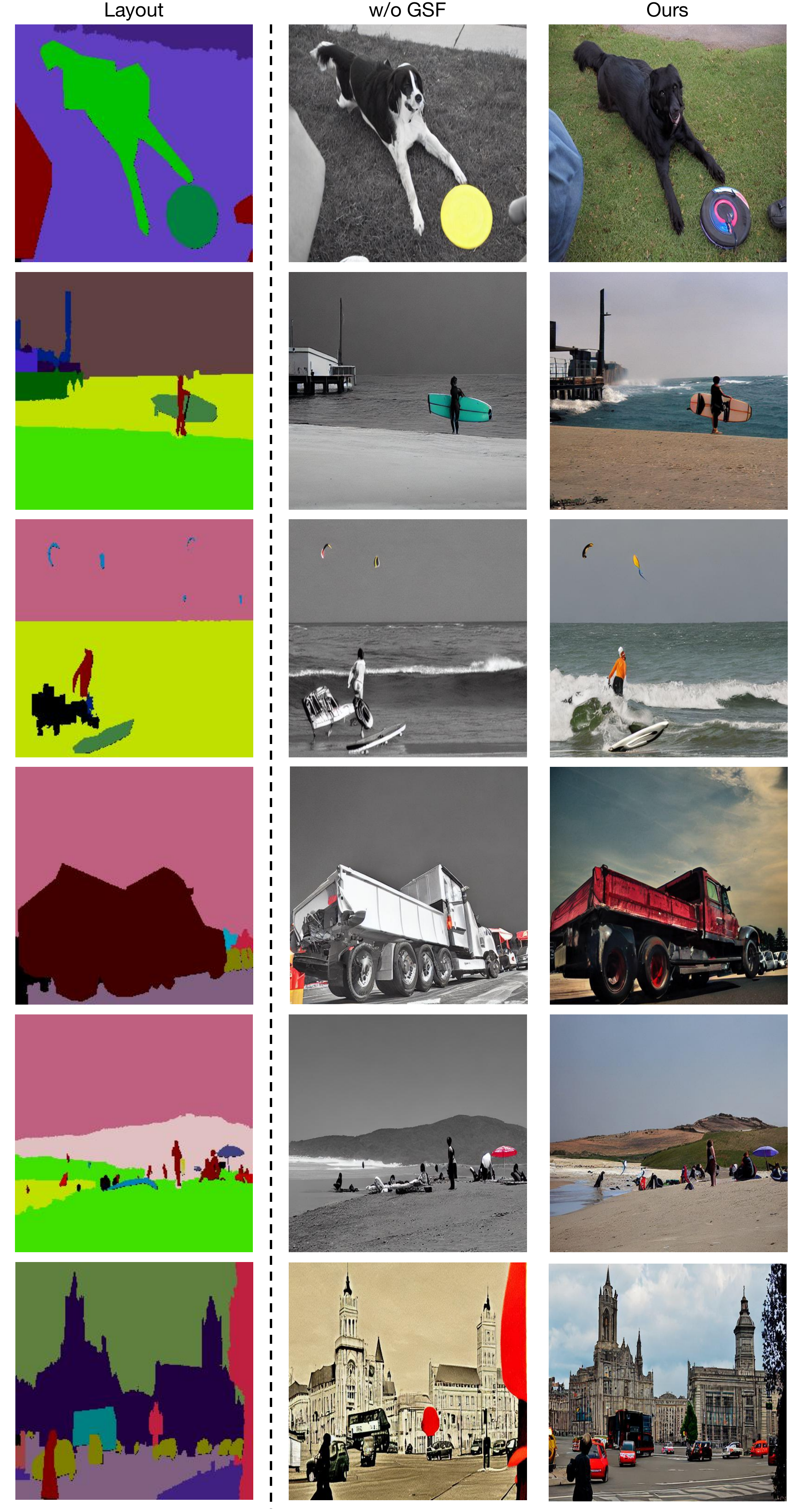}
    \caption{Effectiveness of the proposed GSF. The removal of GSF results in stylistic inconsistency among the generated images. The captions in these cases are set to empty and we omit the grounded text for simplicity.}
    \label{fig:GSF_effective}
\end{figure*}

%% file: figs/ablation_GSF_CA.tex
\begin{figure*}[ht]
    \centering
    \includegraphics[width=1\linewidth]{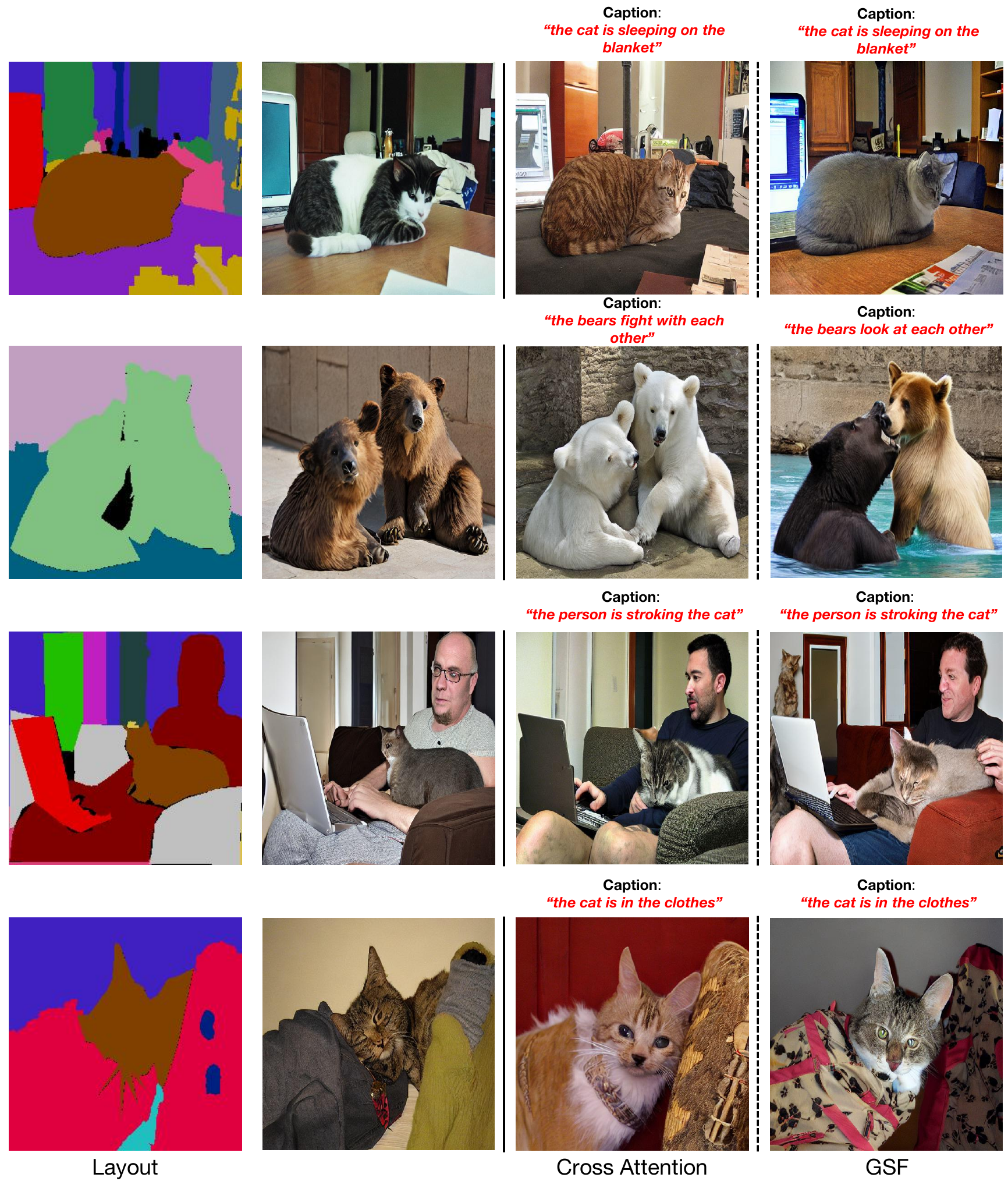}
    \caption{Qualitative results against the method of directly concatenating the caption and grounded text to integrate semantic information into the generation process. We omit the grounded text for simplicity.}
    \label{fig:ablation_GSF_CA}
\end{figure*}

%% file: figs/more_coco.tex
\begin{figure*}[ht]
    \centering
    \includegraphics[width=\linewidth]{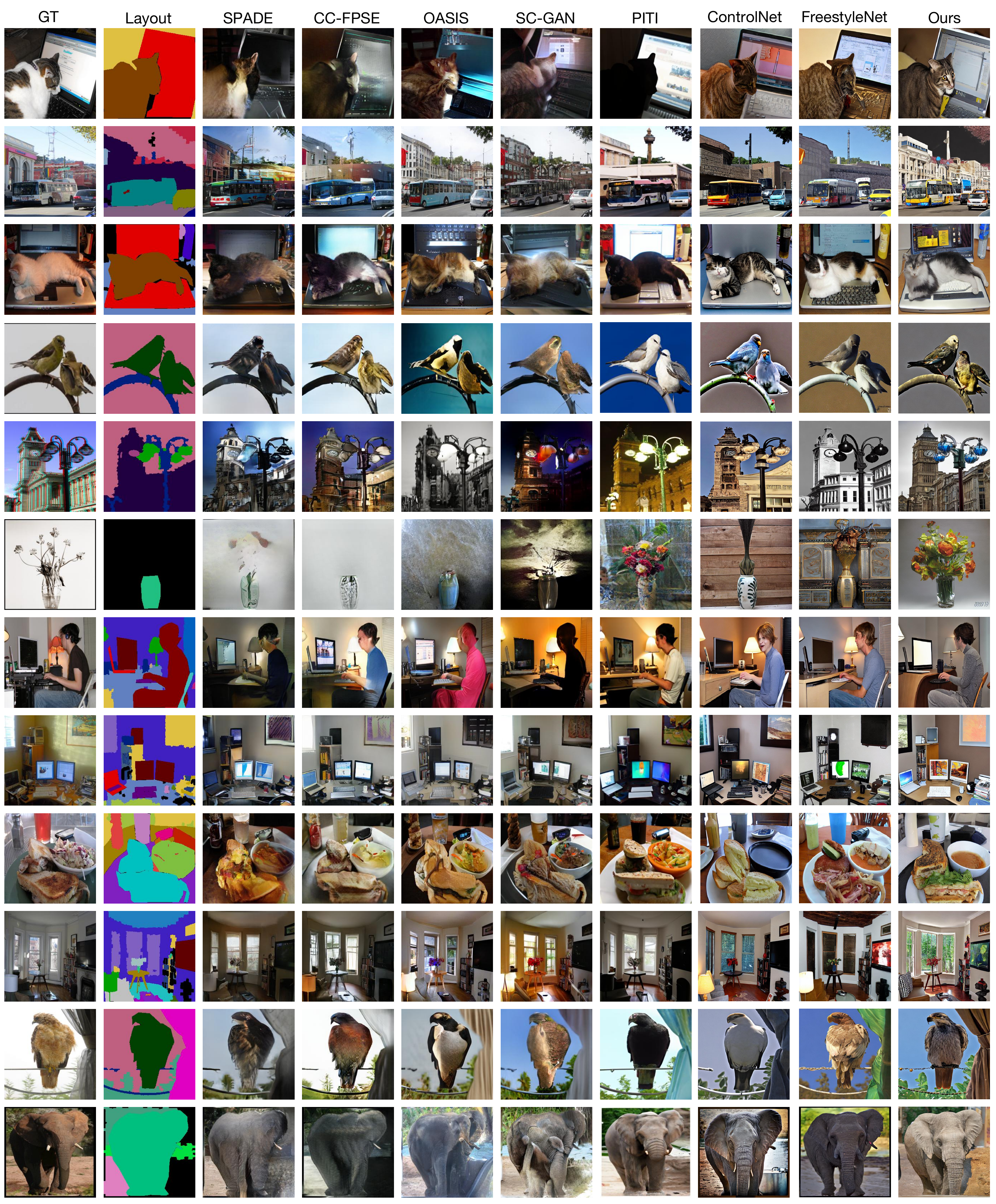}
    \caption{More qualitative comparison results with LIS baselines on COCO-stuff. The captions of our model in these cases are set to empty and we omit the grounded text for simplicity.}
    \label{fig:more_coco}
\end{figure*}

%% file: figs/more_ade20k.tex
\begin{figure*}[ht]
    \centering
    \includegraphics[width=1\linewidth]{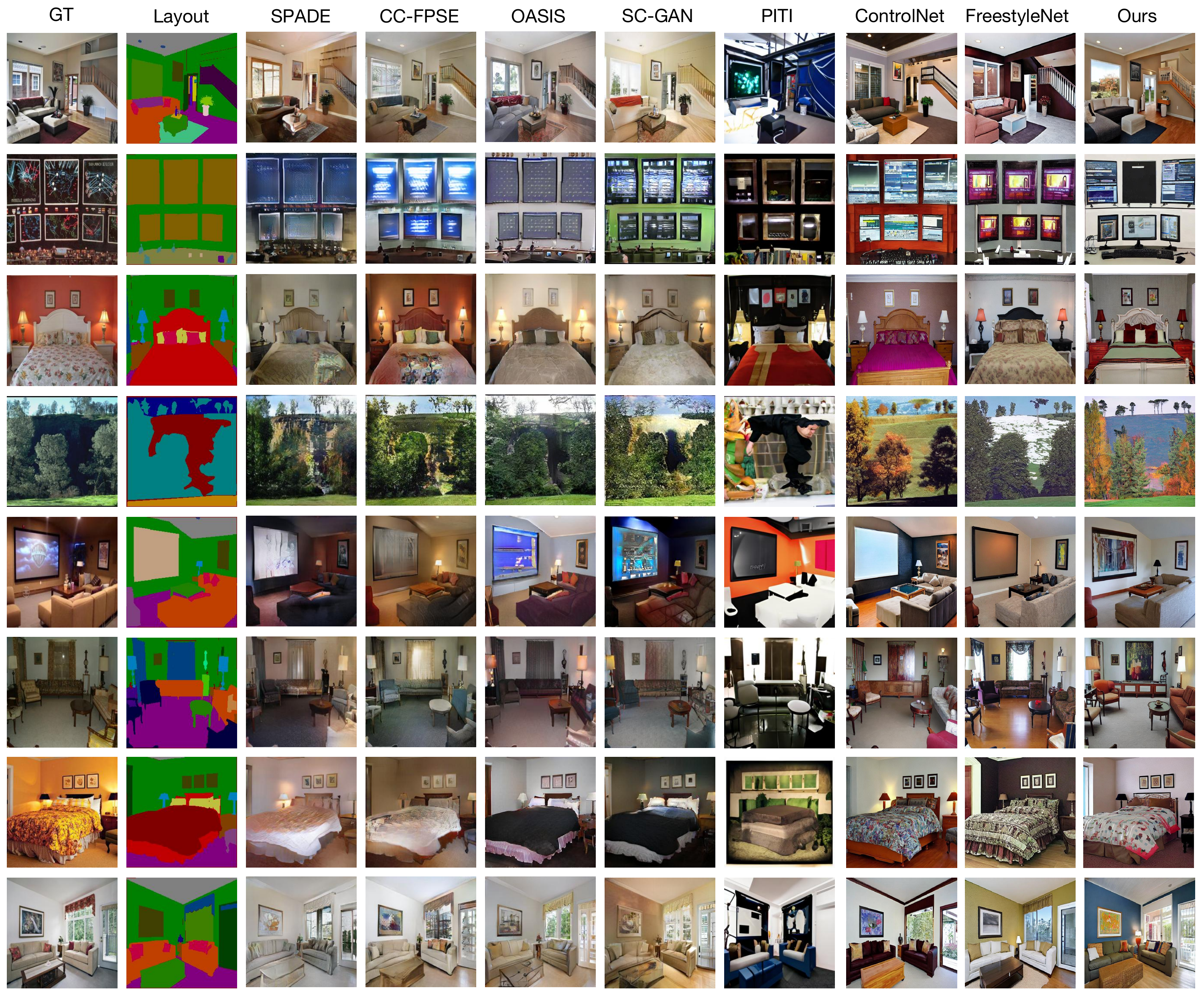}
    \caption{More qualitative comparison results with LIS baselines on ADE20K. The captions of our model in these cases are set to empty and we omit the grounded text for simplicity.}
    \label{fig:more_ade20k}
\end{figure*}